%% file: main.tex
\RequirePackage{iftex}
\ifPDFTeX\pdfoutput=1\fi
\documentclass[a4paper,11pt,onecolumn]{personalarxiv}

\usepackage{fontawesome5}
\usepackage{tikz}
\usetikzlibrary{arrows.meta,positioning,fit,backgrounds,calc}
\renewcommand{\arraystretch}{1.12}
\newcommand{\ind}{\mathbf 1}
\newcommand{\FK}{\operatorname{FK}}
\newcommand{\dist}{\operatorname{dist}}
\newcommand{\clip}{\operatorname{clip}}
\newcolumntype{L}[1]{>{\raggedright\arraybackslash}p{#1}}

\title{\textbf{VRBench: Offline Evaluation of Robotic Manipulation Ability of VLMs}}
\papershorttitle{VRBench}

\renewcommand{\paperauthorgap}{0.9em}
\fancypagestyle{vrbenchfirst}[paperfirst]{%
  
  \fancyhead[L,R]{}%
  \fancyhead[C]{%
    \makebox[\headwidth][l]{%
      \raisebox{-.5\height}{\includegraphics[width=.170\headwidth]{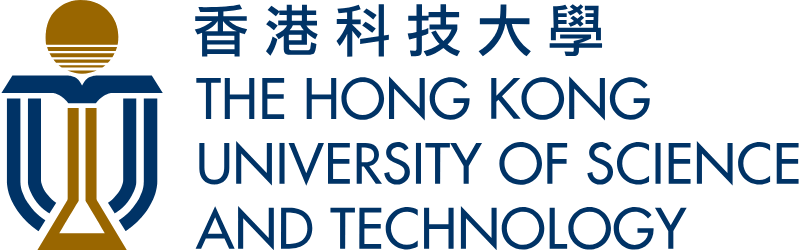}}%
      \hspace{.06\headwidth}%
      \raisebox{-.5\height}{\includegraphics[width=.215\headwidth]{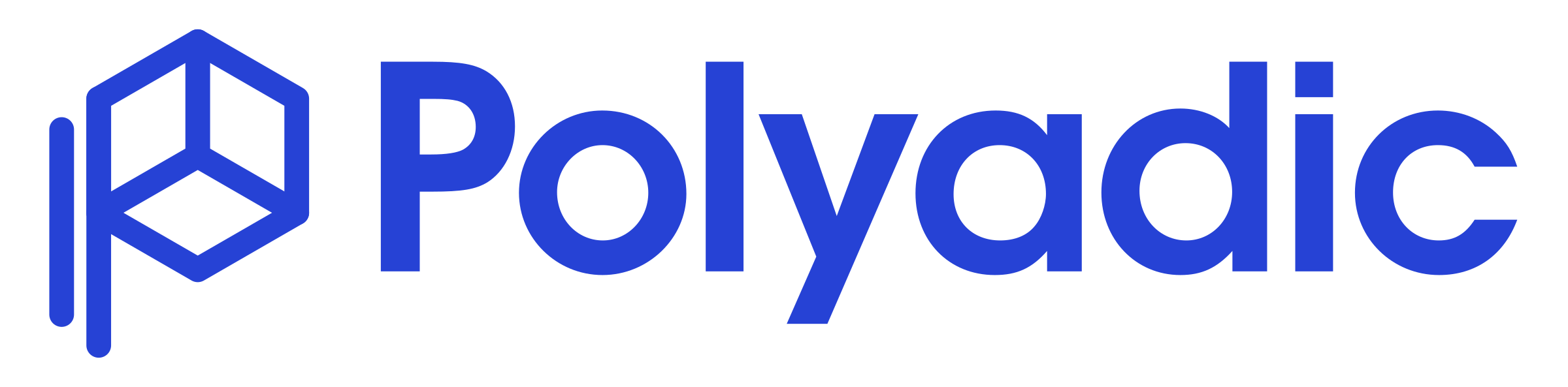}}%
      \hspace{.06\headwidth}%
      \raisebox{-.5\height}{\includegraphics[width=.352\headwidth]{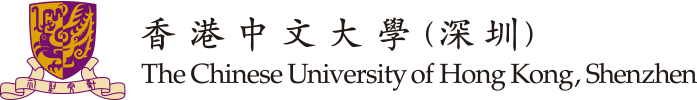}}%
    }%
  }%
}

\renewenvironment{abstract}{%
  \par\begingroup\normalfont\normalsize
  \setlength{\parindent}{0pt}%
  \begin{center}{\large\scshape Abstract}\end{center}%
  \begin{adjustwidth}{0.09\textwidth}{0.09\textwidth}%
  \noindent\ignorespaces
}{%
  \par\end{adjustwidth}\endgroup\smallskip
}

\author[1,2]{Yuxuan Chen}
\author[2,3]{Yihao Shen}
\author[1,2]{Xuantong Liu\thanks{Corresponding author: \href{mailto:xliude@connect.ust.hk}{\texttt{xliude@connect.ust.hk}}}}
\author[4]{Weijian Luo\thanks{Corresponding author: \href{mailto:pkulwj1994@icloud.com}{\texttt{pkulwj1994@icloud.com}}}}
\author[2,3]{Tianyang Hu}
\author[1]{Yuan Yao}

\affil[1]{The Hong Kong University of Science and Technology}
\affil[2]{Polyadic Research Group}
\affil[3]{SAGA Lab, The Chinese University of Hong Kong, Shenzhen}
\affil[4]{Dots Studio, Xiaohongshu}

\begin{document}
\maketitle
\thispagestyle{vrbenchfirst}

\begin{abstract}
Vision-language models (VLMs) increasingly make action-level decisions in robotic manipulation, requiring them to translate task understanding into motion a specific robot can follow.
While closed-loop rollouts test this capability directly at steep API and simulation cost, existing offline benchmarks evaluate only subsets of object and part grounding, interaction selection, spatial goal grounding, action sequencing, and action quality, leaving open how well a VLM integrates them in one complete plan.
We introduce \textbf{VirtualRoboBench (VRBench)}, an offline benchmark designed to comprehensively evaluate a VLM's robotic manipulation capability through a single task: \textbf{action-level manipulation planning}.
From initial calibrated multi-view RGB observations, the robot state, and an instruction, a VLM predicts a 6-DoF end-effector trajectory with gripper and contact events across 50 RoboCasa tasks.
VRBench checks stage regions, poses, events, and ordering, and evaluates action quality via kinematic tracking and collision risk.
Evaluating nine frontier VLMs reveals that robot tracking and collision constraints uncover critical motion failures missed by task-level checks alone, while closed-loop validation confirms strong correlation with task success ($\rho=1.00, r=0.93$) using 97.3\% fewer uncached tokens, 98.1\% lower API costs, 90.8\% less API time, and 89.6\% less total evaluation time per task on average.

\par\smallskip{\centering
  \small
  \textcolor[HTML]{6E40C9}{\faGithub}\enspace
  \textbf{Code:}\enspace
  \href{https://github.com/polyadic-research/VRbench}
       {\nolinkurl{github.com/polyadic-research/VRbench}}\\
  \raisebox{-0.24em}{\includegraphics[height=1.16em]{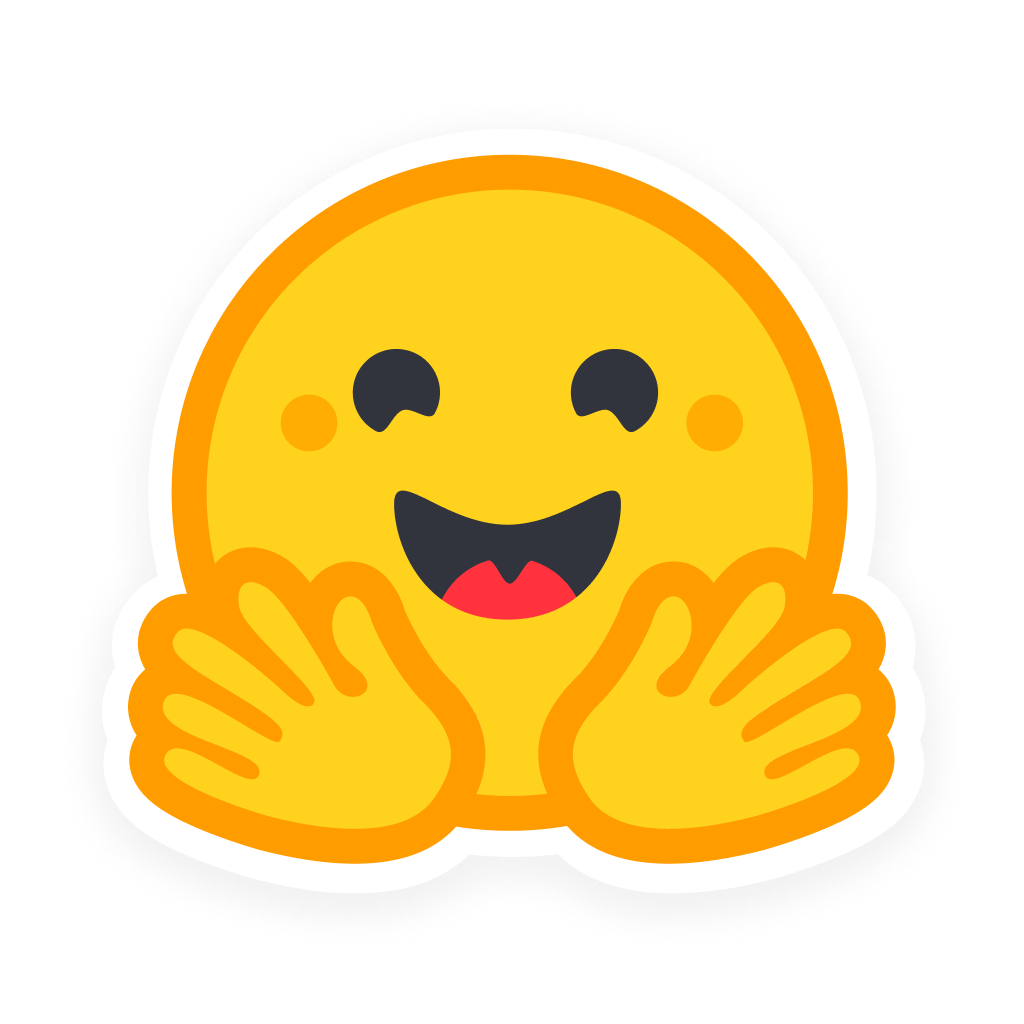}}\enspace
  \textbf{Dataset:}\enspace
  \href{https://huggingface.co/datasets/cyx712/VirtualRoboBench}
       {\nolinkurl{huggingface.co/datasets/cyx712/VirtualRoboBench}}\\
  \textcolor[HTML]{1A73E8}{\faGlobe}\enspace
  \textbf{Project page:}\enspace
  \href{https://yuxuanmath.github.io/VRBench/}
       {\nolinkurl{yuxuanmath.github.io/VRBench/}}\\
  \textcolor[HTML]{0F9D8A}{\faUsers}\enspace
  \textbf{Team page:}\enspace
  \href{https://home.polyadiclabs.com/}
       {\nolinkurl{home.polyadiclabs.com/}}
\par}
\end{abstract}

\noindent\begin{minipage}{\linewidth}
\captionsetup{type=figure}
\input{figures/composite-score}
\caption{VRBench scores (left) and mean runtime and API cost per task for offline and closed-loop evaluation (right).}
\label{fig:score-highlights}
\end{minipage}\par

\section{Introduction}
\label{sec:intro}

As vision-language models (VLMs) are increasingly deployed to guide robotic manipulation \citep{ahn2022saycan,huang2023voxposer,nasiriany2024pivot,liu2024moka,huang2024copa,huang2025rekep,hwang2026hamster,chou2026vflash,tang2026pear,galbot2026astra}, reliably evaluating a VLM's core robotic capability has become a central challenge.
Interactive closed-loop evaluation tests manipulation directly in simulation environments such as RoboCasa \citep{robocasa2024} by pairing a VLM with predefined skills or a low-level action policy \citep{embodiedbench2025,vlaarena2026,shen2026roboharnesszero}.
However, closed-loop rollouts are computationally expensive, require complex execution pipelines, and are difficult to scale across rapidly evolving frontier VLMs.
Offline benchmarks offer a scalable alternative, yet existing suites evaluate robotic capability only in fragments—probing physical affordances or skill ordering through multiple-choice questions and symbolic plans \citep{pacbench2025,zhao2025manipbench,zhang2024vlabench}, or testing isolated 2D/3D spatial pointing and object-level traces \citep{roborefer2025,vabench2025,tracespatial2025}.
Because these abstractions omit 6-DoF end-effector geometry and robot arm kinematics, a VLM can score well offline—for instance, identifying a drawer handle and choosing to pull it—while still proposing an approach pose that cannot grasp the handle or a path that collides with the counter during actual manipulation.

Comprehensively reflecting a VLM's robotic manipulation capability requires evaluating five coupled dimensions together (Table~\ref{tab:benchmark_comparison}).
\emph{Object and part grounding} identifies what to act on, and \emph{interaction selection} determines how to engage it.
\emph{Spatial goal grounding} specifies where the action should lead, and \emph{action sequencing} establishes the order of operations.
Crucially, \emph{action quality} determines whether the target robot arm can accurately track the proposed poses and move safely through the scene without collision.
Existing offline benchmarks cover only subsets of these dimensions, leaving open a key question: can a unified offline benchmark capture all five dimensions and strongly reflect how VLMs perform in closed-loop robotic manipulation?

\suppressfloats[t]
\begin{figure}[!tb]
\centering
\includegraphics[width=\linewidth]{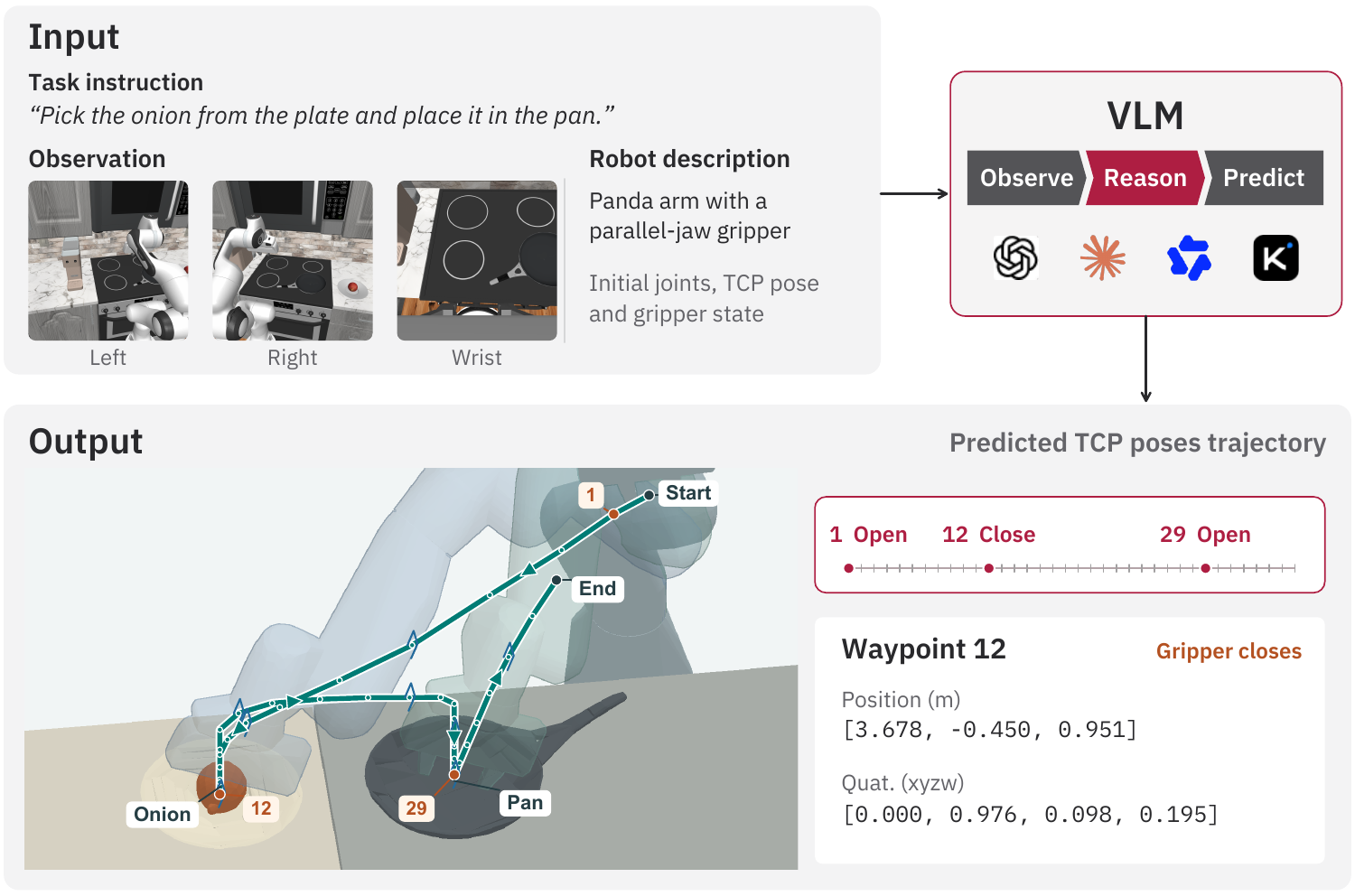}
\caption{\textbf{Action-level manipulation planning in VRBench.}
Given three initial camera views, the task instruction, and the robot's initial state, a VLM predicts a 6-DoF tool-center-point (TCP) trajectory to move an onion from a plate to a pan.
Arrows indicate the direction of motion from the initial TCP position (Start) to the final predicted waypoint (End).
The numbered waypoints mark gripper events.
The translucent arms show the reconstructed poses for grasping the onion (12) and releasing it over the pan (29).
The inset shows the TCP position and orientation at waypoint~12, where the plan calls for the gripper to close.}
\label{fig:action-plan}
\end{figure}

We introduce \textbf{VirtualRoboBench (VRBench)}, an offline benchmark designed to comprehensively evaluate a VLM's robotic manipulation capabilities through a single unified task: \emph{action-level manipulation planning}.
Given initial calibrated multi-view RGB observations, a task instruction, and the robot's initial state, the VLM predicts a complete task-length sequence of 6-DoF tool-center-point (TCP) poses with gripper and contact events (Figure~\ref{fig:action-plan}).
Expressing an entire manipulation task in this format requires the model to integrate all five capabilities in one coherent plan—for example, when moving a pan into a sink, selecting a valid grasp pose on the handle, closing the gripper before transport, targeting the sink for release, and guiding the arm along a collision-free path.

VRBench instantiates this evaluation across 50 household manipulation tasks from RoboCasa \citep{robocasa2024} using a modular two-branch evaluator.
For each task stage, the task evaluator checks where and at what orientation an interaction is proposed, whether the gripper or contact event is appropriate, and whether prerequisite stages have been completed in order.
In parallel, the robot evaluator reconstructs the arm's joint trajectory from its initial configuration via inverse kinematics, measuring both TCP tracking accuracy and arm--scene collision risk.
Combining task progress with robot compatibility yields a holistic offline measure of manipulation capability without requiring simulator rollouts.

Evaluating nine frontier VLMs on VRBench reveals substantial capability gaps and shows that robot tracking and collision constraints uncover critical planning failures missed by task-level checks alone (Figure~\ref{fig:score-highlights}).
To test whether VRBench reflects actual manipulation performance, we compare offline scores against closed-loop execution in RoboCasa using a shared downstream policy so that all VLMs are judged under identical control capabilities.
Across all six models evaluated in closed-loop simulation, the ranking by VRBench composite score matches the ranking by closed-loop task success rate (Spearman's $\rho = 1.00$, Pearson's $r = 0.93$), and higher offline scores predict the winner in 62\% of same-task comparisons with different closed-loop outcomes.
At the same time, single-pass offline evaluation reduces uncached VLM token consumption by 97.3\% ($37\times$ fewer tokens), API waiting time by 90.8\% ($11\times$ faster), and local per-task runtime by $10\times$ compared with closed-loop rollouts.
These results demonstrate that VRBench provides a lightweight, reproducible offline evaluation that closely tracks real simulator manipulation performance at a fraction of the cost.
\par\addvspace{0.8\baselineskip}
\noindent\textbf{Contributions.}\par\nopagebreak
\begin{itemize}[topsep=1pt,partopsep=0pt]
    \item \textbf{Comprehensive offline benchmark for VLM robotic capability.}
    VRBench formulates action-level manipulation planning across 50 RoboCasa tasks to jointly evaluate object and part grounding, interaction selection, spatial goal grounding, action sequencing, and action quality from a single prediction.
    \item \textbf{Modular task-and-robot evaluator.}
    Stage-graph rules assess goal satisfaction and interaction ordering, while inverse-kinematics arm reconstruction evaluates tracking accuracy and collision risk without physics rollouts.
    \item \textbf{Execution-aligned evaluation at low cost.}
    Across the models evaluated in closed loop, offline scores correlate with simulation success ($\rho=1.00$, $r=0.93$). On the paired cost comparison, VRBench used 97.3\% fewer uncached tokens and reduced average evaluation time per task by 89.6\%.
\end{itemize}

\section{Related Work}

\label{sec:related}
\paragraph{VLMs for Manipulation and Closed-Loop Evaluation.}
Recent embodied frameworks increasingly use VLMs not only to sequence high-level skills \citep{ahn2022saycan,huang2023voxposer}, but also to generate 2D/3D keypoint paths \citep{nasiriany2024pivot,liu2024moka,hwang2026hamster,chou2026vflash}, 6-DoF end-effector poses and constraints \citep{huang2024copa,huang2025rekep,tang2026pear}, or online action corrections \citep{galbot2026astra}.
To evaluate manipulation in simulation, environments such as RoboCasa \citep{robocasa2024} and policy benchmarks such as VLA-Arena \citep{vlaarena2026} test end-to-end visuomotor policies, whereas VLM-oriented frameworks such as EmbodiedBench \citep{embodiedbench2025} and RoboHarness-Zero \citep{shen2026roboharnesszero} couple a VLM planner with low-level skills or pretrained policies to measure closed-loop task success.
While closed-loop execution directly reflects end-to-end performance, repeated rollouts are resource-intensive and entangle the VLM's planning quality with the downstream controller.
VRBench complements closed-loop evaluation by turning RoboCasa tasks into a rollout-free offline benchmark that still strongly correlates with closed-loop manipulation success.

\begin{table}[h]
\centering\footnotesize
\setlength{\tabcolsep}{5pt}
\renewcommand{\arraystretch}{1.16}
\begin{tabular}{@{}L{0.36\linewidth}L{\dimexpr0.64\linewidth-2\tabcolsep\relax}@{}}
\toprule
\textbf{Benchmark} & \textbf{Evaluation dimensions}\\
\midrule
RefSpatial-Bench \citep{roborefer2025} & Object and part grounding, Spatial goal grounding\\
PAC Bench \citep{pacbench2025} & Interaction selection\\
ManipBench \citep{zhao2025manipbench} & Object and part grounding, Interaction selection, Spatial goal grounding, Action sequencing\\
VLABench \citep{zhang2024vlabench} & Object and part grounding, Interaction selection, Action sequencing\\
VABench \citep{vabench2025} & Spatial goal grounding, Action quality (2D trace accuracy)\\
GraspNet-1Billion \citep{fang2020graspnet} & Action quality (grasp pose quality)\\
DroidSpatial-Bench \citep{hwang2026hamster} & Object and part grounding, Spatial goal grounding\\
TraceSpatial-Bench \citep{tracespatial2025} & Object and part grounding, Spatial goal grounding, Action quality (object--scene collision avoidance)\\
\midrule
\textbf{VRBench (Ours)} & \textbf{Object and part grounding, Interaction selection, Spatial goal grounding, Action sequencing, Action quality} (kinematic tracking accuracy and arm--scene collision avoidance)\\
\bottomrule
\end{tabular}
\caption{\textbf{Evaluation dimensions in offline robotic manipulation benchmarks.}
VRBench assesses all five dimensions jointly from a single task-length sequence of 6-DoF end-effector poses with gripper and contact events.}
\label{tab:benchmark_comparison}
\end{table}

\paragraph{Offline Benchmarks for Robotic Manipulation.}
Existing offline benchmarks evaluate different subsets of manipulation decisions without physics rollouts (Table~\ref{tab:benchmark_comparison}).
Several suites focus on semantic and spatial reasoning before motion: RefSpatial-Bench tests 2D pointing to objects or goal locations specified by spatial relations \citep{roborefer2025}, while PAC Bench probes physical properties, affordances, and interaction constraints \citep{pacbench2025}.
In offline planning tracks, VLABench evaluates symbolic skill selection and ordering \citep{zhang2024vlabench}, and ManipBench uses multiple-choice questions to test reasoning about low-level actions and relates those scores to real-robot action selection \citep{zhao2025manipbench}.
Although a correct answer or skill plan establishes what to act on and in what order, it leaves continuous 6-DoF poses and the robot's motion between interactions unspecified.
Other offline benchmarks evaluate spatial geometry and trajectories, but do not jointly model multi-stage task logic and articulated arm kinematics.
Classical grasp benchmarks such as GraspNet-1Billion score isolated 6-DoF grasp poses via gripper collision and analytic wrench metrics without language-conditioned task planning \citep{fang2020graspnet}.
Among VLM benchmarks, VABench evaluates targets and 2D motion traces in the image plane \citep{vabench2025}, DroidSpatial-Bench measures 3D endpoint errors at grasp and placement locations without scoring intermediate motion \citep{hwang2026hamster}, and TraceSpatial-Bench checks whether a disembodied moving object avoids scene obstacles along a 3D path \citep{tracespatial2025}.
In contrast, VRBench elicits a complete 6-DoF end-effector trajectory with gripper and contact events, jointly evaluating multi-stage task goals alongside kinematic tracking accuracy and collision risk on the reconstructed robot arm.

\section{VRBench Design}
\label{sec:benchmark}
\begin{figure}[!tb]
\centering
\includegraphics[width=\linewidth]{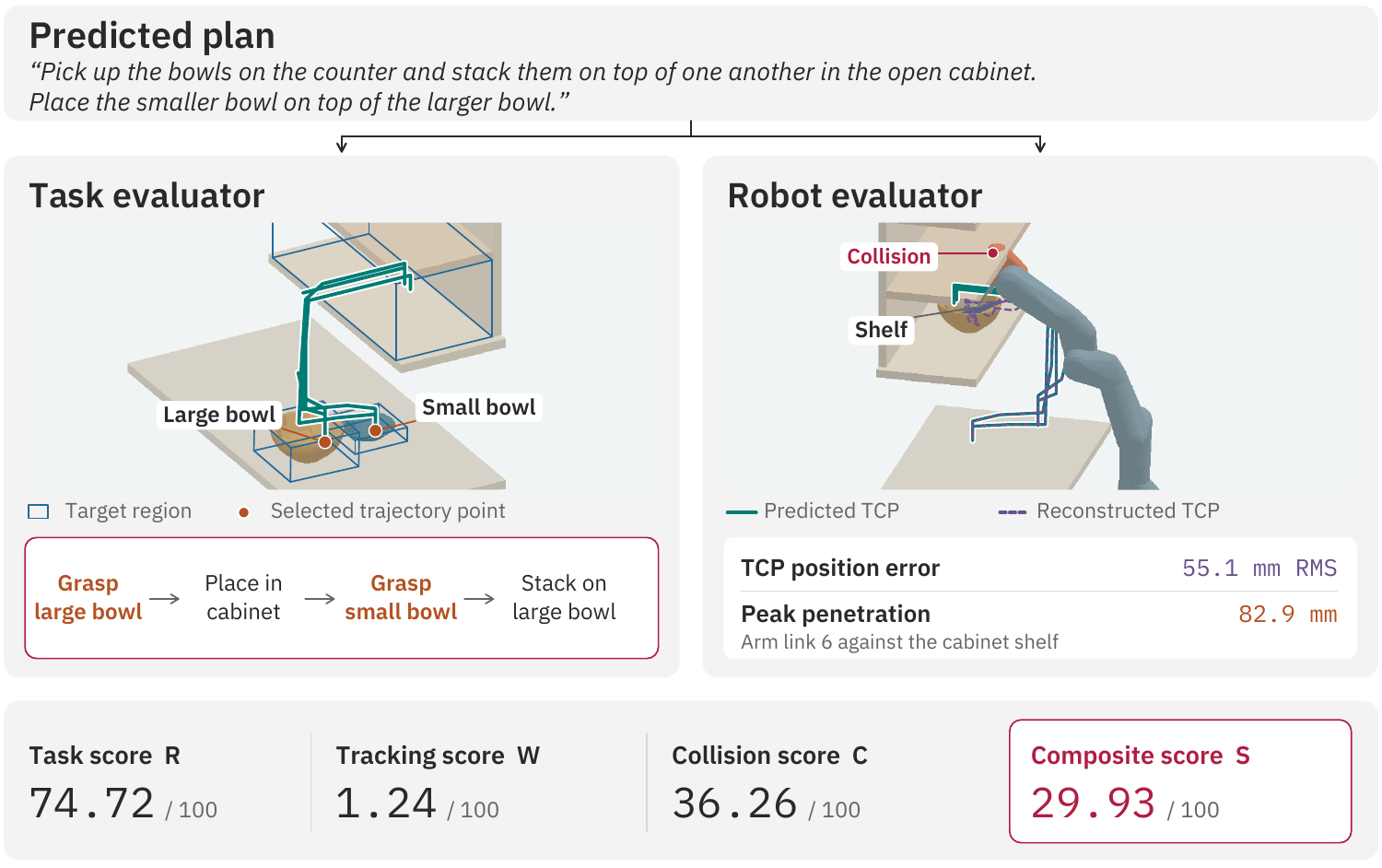}
\caption{\textbf{VRBench score.} The plan satisfies the requirements for grasping the larger bowl and placing it in the cabinet. The grasp point for the smaller bowl falls just outside its target region, earning a partial score and leaving the final stacking stage at zero. The right panel highlights the reconstructed arm link where it penetrates the cabinet shelf most deeply. The bottom row reports the task score $R$, tracking score $W$, collision score $C$, and composite score $S$ computed from the stage scores.}
\label{fig:overview}
\end{figure}

VRBench is a lightweight, rollout-free framework for evaluating action-level manipulation planning in general-purpose VLMs.
Figure~\ref{fig:action-plan} illustrates the model input and output, and Figure~\ref{fig:overview} shows how the task and robot evaluators score a predicted trajectory.
We introduce the general framework and its RoboCasa instantiation in this section, followed by the scoring methods in Section~\ref{sec:metrics}.

\subsection{Framework}
\label{sec:architecture}

VRBench decouples a robot-agnostic evaluation engine from pluggable task suites so that any off-the-shelf VLM can be evaluated rapidly without physics simulation or robot-specific fine-tuning.
The framework defines a common prediction format and scoring workflow, while the task suite supplies task goals, interaction requirements, and robot specifications.

Each evaluation starts from initial observations, a task instruction, and the initial robot state.
Based on visual and state inputs alone, the VLM predicts a complete trajectory of $K$ 6-DoF TCP poses annotated with manipulation events.
The task evaluator checks whether the planned poses and events satisfy the stage goals in the required order, while the robot evaluator measures kinematic tracking accuracy and collision risk along the reconstructed arm motion.
Combining these stage-level evaluations produces the composite score, and supporting different tasks or robots requires changing only the task rules and robot settings.

\subsection{Input and Output}
\label{sec:interface}

For each episode, the VLM receives initial multi-view RGB observations $I^0$, camera calibration $\mathcal K$, a natural-language instruction $L$, and the initial robot state $s^0=(T^0,q^0,g^0)$ comprising the initial TCP pose, joint configuration, and gripper state.
Through the prompt, the input also provides the robot information, coordinate conventions, waypoint count $K$, and allowed manipulation events needed to specify the trajectory.

The VLM outputs its plan in a single pass as $K$ ordered waypoints:
\begin{equation}
\hat\tau=
\{(\hat p_k,\hat Q_k,\hat g_k,\hat c_k)\}_{k=1}^{K},
\label{eq:trajectory}
\end{equation}

where $\hat p_k\in\mathbb R^3$ is the TCP position in meters and $\hat Q_k\in\mathrm{SO}(3)$ is its orientation represented as a unit quaternion in a common world coordinate frame, and $\hat g_k$ and $\hat c_k$ denote the predicted gripper event and contact event at waypoint $k$, respectively.
Because these manipulation events are defined by the task configuration rather than tied to a specific low-level controller, the interface applies directly to any general-purpose VLM.
All models are evaluated under identical settings for each episode.

\subsection{RoboCasa Task Suite}
\label{sec:construction}

We instantiate VRBench on 50 selected manipulation tasks from RoboCasa \citep{robocasa2024}. These tasks involve household objects and fixtures, providing different manipulation goals and interaction requirements. The selected tasks form the basis of our evaluation suite.

Each task provides the information needed by the framework, including initial observations and instructions, task goals and stage requirements, trajectory settings, and robot parameters. The task package also includes the geometry and configuration data needed for evaluation.

To prepare an evaluation episode, we specify a fixed initial scene, object placements, camera views, and initial robot state, and represent valid task execution as a set of stages and their ordering constraints. For example, transferring an object may involve reaching the object, grasping it, moving it, and releasing it at the target. 
Because these annotations specify valid stage outcomes rather than a single expert trajectory, distinct predicted paths score equally well whenever they meet the same spatial and interaction requirements (Section~\ref{sec:task-score}).

We focus on tasks whose key goals and interactions can be assessed geometrically from the predicted trajectory and scene state without contact-physics simulation.
For tasks that alter the scene, such as moving an object or opening a door, the task configuration specifies either fixed scene geometry or task-stage-dependent updates for goal and collision checks.
We validate all 50 task configurations using both valid example trajectories and controlled invalid trajectories—covering incorrect targets, missing events, ordering violations, and collisions—to ensure that the evaluator reliably separates valid plans from common planning errors.

\section{Evaluation}
\label{sec:metrics}

VRBench evaluates predicted trajectories from two complementary perspectives that jointly cover all five evaluation dimensions.
The task evaluator assesses object and part grounding, interaction selection, spatial goal grounding, and action sequencing, producing a task score $R$.
The robot evaluator assesses action quality through kinematic tracking accuracy and arm--scene collision risk, yielding the tracking score $W$ and collision score $C$. 
To compute the composite score $S$, we combine the task, tracking, and collision scores within each stage and take a weighted average across stages. 
The following sections describe the evaluation process and scoring methods, with detailed robot scoring formulas provided in Appendix~\ref{app:robot-scoring}.

\subsection{Task Evaluator}
\label{sec:task-score}

We divide each task into stages—such as grasping an object, transporting it, and placing it at a destination—and represent their ordering constraints as a directed acyclic graph $\mathcal D=(\mathcal V,\mathcal E)$.
Each stage $i$ specifies a target region $\Omega_i$ (reflecting object or part grounding for manipulation stages and spatial goal grounding for placement stages) and may require specific TCP poses or manipulation events $\mathcal H_i$, such as closing or opening the gripper.
Prerequisite stages must be completed before dependent stages, whereas independent stages may occur in any order.

We first measure how well the predicted trajectory $\hat{\tau}$ meets each stage's requirements. 
For a distance-based stage, let $d_i$ be the smallest residual over trajectory samples eligible for that interaction. Its soft score is

\begin{equation}
d_i=\dist_i(\hat\tau,\Omega_i,\mathcal H_i),\qquad
r_i=100\exp(-d_i/\sigma_i),
\label{eq:region}
\end{equation}

where $\sigma_i$ is expressed in the same units as $d_i$. A stage requiring an event receives zero if there is no eligible event. Some stages measure progress directly, such as the spread of contacts while wiping. Appendix~\ref{app:task-scoring} gives the distance, completion and weighting rules for the RoboCasa suite.

For tasks involving articulated objects, stage-specific trajectory evidence may include reaching a handle region or moving along the opening direction of a drawer.
These checks provide geometric evidence of the intended operation.

A high stage score alone does not mean that the stage is completed. We set $h_i=1$ when the stage's geometric and operation conditions are met, including an orientation check where specified, and $h_i=0$ otherwise. We also check whether the required earlier stages have been completed. 
The gated score for stage $i$ is

\begin{equation}
a_i=\prod_{j\in\operatorname{pa}(i)}a_jh_j,\qquad
\tilde r_i=a_ir_i,
\label{eq:gate}
\end{equation}

where $\operatorname{pa}(i)$ denotes the stages that must precede stage $i$, and $a_i=1$ for stages without prerequisites.
A stage is evaluated only after its prerequisites have been satisfied, on the trajectory segment following their completion waypoints.
If a required earlier stage is not completed, its dependent stages score zero, while independent stages are evaluated separately.
Reaching the final target therefore contributes to the task score only if all required preceding stages are completed.

Finally, the task score $R$ is the weighted average of all stage scores:

\begin{equation}
R=\frac{\sum_i w_i\tilde r_i}{\sum_i w_i},
\qquad w_i>0,
\label{eq:R}
\end{equation}

where $w_i$ is the weight of stage $i$. The 50 RoboCasa tasks weight their stages equally.

\subsection{Robot Evaluator}
\label{sec:robot}

The robot evaluator checks whether the predicted TCP trajectory can be followed by the robot and whether the resulting motion collides with the environment.
It consists of joint reconstruction, kinematic tracking, and collision checking.

We first use inverse kinematics (IK) to reconstruct a joint trajectory from the given initial robot configuration without obstacle rerouting (see Appendix~\ref{app:joint-reconstruction}).
We then compare the reconstructed TCP poses with the predicted poses, measuring position errors, orientation errors, and large joint changes.
These measurements produce a tracking score $W$ on a scale of 0 to 100.

Next, we check collisions between the reconstructed robot arm and the environment.
We interpolate the joint trajectory to examine intermediate robot configurations, rather than checking only the predicted waypoints.
The collision score $C$, also on a scale of 0 to 100, considers the depth and frequency of geometric penetration, with a geometric tolerance and task-specific contact exclusions.

Both scores are computed for the complete trajectory and for individual task stages. For each stage $i$, we obtain a tracking score $W_i$ and a collision score $C_i$ over its corresponding motion interval. These scores are combined with the task-stage scores in Section~\ref{sec:composite} to calculate the composite score, with detailed tracking and collision formulas given in Appendix~\ref{app:robot-scoring}.

\subsection{Composite Score}
\label{sec:composite}

We combine task progress and robot compatibility at each task stage. Let $\tilde r_i$ be the gated task-stage score from Eq.~\eqref{eq:gate}, and let $C_i$ and $W_i$ be the corresponding collision and tracking scores from Section~\ref{sec:robot}.

Let $I_i$ be the motion interval for stage $i$ and $p_{I_i}^{\max}$ its largest sampled excess penetration after subtracting the geometric tolerance (Appendix~\ref{app:robot-scoring}). We apply score cutoffs $\theta_C$ and $\theta_W$ and a depth cutoff $\theta_p$ when computing the stage factors.
\[
\bar C_i=C_i\,\ind[C_i\ge\theta_C]\,\ind[p_{I_i}^{\max}<\theta_p],
\qquad
\bar W_i=W_i\,\ind[W_i\ge\theta_W].
\]
A single sample at or above the depth cutoff therefore removes the robot-dependent contribution for that stage, even if its raw collision score passes the score cutoff.

The composite score is defined as
\begin{equation}
S=
\frac{
\sum_i w_i\tilde r_i
\left[
\alpha+(1-\alpha)
\frac{\bar C_i}{100}
\frac{\bar W_i}{100}
\right]
}{
\sum_i w_i
},
\label{eq:overall-score}
\end{equation}
where $w_i>0$ is the weight of stage $i$, and $\alpha\in[0,1]$ controls the balance between task progress and robot compatibility. The numerical settings are given in Section~\ref{sec:setup}.

A stage with zero gated task score contributes nothing to the composite score. For stages with positive task scores, higher tracking and collision scores increase the contribution when the stage passes both robot checks. If either check fails, the robot-dependent contribution is removed, while the task component is retained.

This design rewards plans that satisfy task requirements while accounting for robot motion constraints. 
Note that $W$ and $C$ evaluate motion feasibility conditioned on the proposed path: a conservative or nearly stationary trajectory can achieve high tracking and collision scores while failing to reach target objects.
Multiplying the robot factor by $\tilde r_i$ in Eq.~\eqref{eq:overall-score} prevents such passive predictions from earning composite score unless they also make genuine task progress.
We also report the task score $R$ and complete-trajectory tracking and collision scores $W$ and $C$ separately for diagnostic analysis.
For each model, we average scores across all 50 tasks with equal weight, using the numerical parameter settings given in Section~\ref{sec:setup}.

\section{Experiments}
\label{sec:experiments}

\subsection{Experimental Setup}
\label{sec:setup}

We evaluate nine VLMs on the same 50 RoboCasa tasks, with 25 atomic tasks and 25 composite tasks. For each task, every model receives the same instruction, initial robot state, camera calibration, and three $1024\!\times\!1024$ RGB images in the same order. We query each model through its API with tools disabled. We use medium reasoning effort for all models except Kimi K3.1 Pro and DeepSeek-V4.1-Flash, whose evaluation interfaces do not offer a medium setting. For these two models, we use high reasoning effort. Each prediction specifies 36 world-frame TCP poses and their gripper and contact events. When a quaternion fails the unit-norm check, we normalize it and validate the complete prediction.

The task and robot evaluators use the same regions, stage dependencies, robot geometry, and reconstruction settings for all models. We assign each stage weight $w_i=1$. For explicit distance checks, we choose $\sigma_i\in\{10,50\}$\,mm according to the size of the object or target part. Completion requires zero residual, with a numerical tolerance of $10^{-9}$\,m. Each task specifies its accepted regions and operation-specific conditions (Appendix~\ref{app:task-scoring}). Appendix~\ref{app:joint-reconstruction} gives the IK tolerances and search budgets.

In Eq.~\eqref{eq:overall-score}, we set $\alpha=0.10$, use score cutoffs $\theta_C=30$ and $\theta_W=40$ on the 0--100 scale, and set the excess penetration cutoff to $\theta_p=50$\,mm. A stage therefore retains 10\% of its task-stage score if either robot score falls below its cutoff or peak penetration reaches 50\,mm beyond the geometric tolerance. Otherwise the remaining 90\% is scaled by the product of the two coefficients. Tracking uses error scales $\lambda_p=10$\,mm and $\lambda_r=5^\circ$ and joint-change thresholds $\tau_{95}=30^\circ$ and $\tau_{\max}=90^\circ$, with $\beta_J=0.5$. Collision checking uses a maximum joint-space increment $\Delta_{\mathrm{col}}=0.04$\,rad, a penetration tolerance $m_{\mathrm{col}}=10$\,mm, and a decay scale $\lambda_c=10$\,mm. We compute the composite score stage by stage within each task, then average over tasks with equal weight.

\subsection{Offline Planning Performance}
\label{sec:rq1}

Claude Fable 5.2 achieves the highest mean composite score across the 50 tasks (54.72), followed by GPT-6 Astra (53.60) and GPT-6.1 Sol (50.82). Kimi ranks fourth at 46.34, and scores across all nine models span 6.48 to 54.72 (Table~\ref{tab:offline-ranking}). Fable's lead over Astra is small relative to variation across tasks. A paired bootstrap with 20,000 draws stratified by task type gives a 95\% percentile interval of $[-7.45,\,9.89]$ for their 1.12-point difference.

Astra earns the highest task score ($R=86.06$). The task score jointly assesses object and part grounding, interaction selection, spatial goal grounding, and action sequencing. Fable leads on tracking ($W=74.42$) and collision ($C=76.06$), the two measures of action quality. Its advantage on these measures helps explain why it ranks first overall.

\begin{table}[!htb]
\centering\small
\setlength{\tabcolsep}{4.5pt}
\renewcommand{\arraystretch}{1.12}
\begin{tabular}{@{}llrrrr@{}}
\toprule
Model & Reasoning effort & Task $R$ & Tracking $W$ & Collision $C$ & Composite $S$ \\
\midrule
Claude Fable 5.2 & Medium & 81.50 & \textbf{74.42} & \textbf{76.06} & \textbf{54.72} \\
GPT-6 Astra & Medium & \textbf{86.06} & 65.14 & 74.02 & 53.60 \\
GPT-6.1 Sol & Medium & 85.04 & 59.64 & 70.77 & 50.82 \\
Kimi K3.1 Pro & High & 77.54 & 61.26 & 70.22 & 46.34 \\
GPT-6.1 Terra & Medium & 69.98 & 56.69 & 69.11 & 37.62 \\
Qwen3.8-Max & Medium & 63.66 & 56.22 & 66.38 & 34.56 \\
Gemini 4 Argon & Medium & 50.46 & 69.07 & 68.92 & 31.88 \\
DeepSeek-V4.1-Flash & High & 43.89 & 51.40 & 65.08 & 23.29 \\
GPT-6 Luna & Medium & 16.83 & 44.98 & 63.59 & 6.48 \\
\bottomrule
\end{tabular}
\caption{\textbf{Offline scores on 50 RoboCasa tasks.} Scores are means across tasks on a 0--100 scale. The composite score $S$ combines the task, tracking, and collision scores at each stage. Bold marks the highest value in each column.}
\label{tab:offline-ranking}
\end{table}

\begingroup\clubpenalty=10000
GPT-6 Luna ranks last overall ($S=6.48$), driven primarily by severe errors in metric 3D grounding and pose feasibility rather than missing event syntax.
Its task score averages 16.83 and falls below 10 on 29 of 50 tasks, while its tracking score is also the lowest among all models ($W=44.98$).
Although Luna often emits plausible gripper closure events, its predicted world-frame coordinates drift far from target objects: when asked to open a drawer and retrieve a measuring cup, it places both gripper closures more than 50\,cm from the nearest handle-region center, leaving the drawer closed and blocking the retrieval and placement stages at zero (Figure~\ref{fig:case-luna-drawer}).
Even when it approaches closer—as on the onion-transfer task, where the grasp earns 12.16 out of 100—inconsistent 6-DoF poses yield poor arm tracking ($W_i=16.41$) and reduce the composite score to 0.61 (Figure~\ref{fig:case-onion}).\par\endgroup

The spatula task shows how a plan can succeed at interaction selection yet fall short on object and part grounding (Fig.~\ref{fig:case-spatula}). Argon calls for the gripper to close before pressing the chicken, but places the grasp 14.9\,cm from the spatula handle's accepted region. The grasp earns 5.05 points, leaving the pressing stage blocked. The arm tracks the requested poses closely ($W=99.42$), yet the task score is only 2.53.

The bowl-stacking task shows how spatial goal grounding and action sequencing enter the score (Fig.~\ref{fig:overview}). Astra specifies a release of the larger bowl inside the cabinet's accepted placement region, earning 100 for that stage. Its subsequent grasp of the smaller bowl falls just outside the target region. The grasp still earns 98.87 points for proximity, but stacking requires a completed grasp, so the final stage earns zero. The ordering check thus depends on completing the prerequisite interaction as well as placing the events in sequence.

\begin{figure}[!htb]
\centering
\begin{tikzpicture}[x=.147cm,y=.50cm,font=\footnotesize]
\definecolor{atomMark}{HTML}{28746E}
\definecolor{compMark}{HTML}{AB563A}
\definecolor{plotGuide}{HTML}{D5D8D8}
\foreach \x in {0,20,40,60} {
  \draw[plotGuide,densely dotted] (\x,.35)--(\x,-8.48);
  \node[anchor=north,text=black!70] at (\x,-8.50) {\x};}
\fill[atomMark] (9,1.05) circle[radius=2.3pt];
\node[anchor=west] at (10,1.05) {Atomic tasks};
\draw[compMark,line width=1pt,fill=white] (37,1.05) circle[radius=2pt];
\node[anchor=west] at (38,1.05) {Composite tasks};
\def\grouprow#1#2#3#4{%
  \node[anchor=east] at (-4,#2) {#1};
  \draw[plotGuide,line width=1.4pt] (#3,#2)--(#4,#2);
  \fill[atomMark] (#3,#2) circle[radius=2.3pt];
  \draw[compMark,line width=1pt,fill=white] (#4,#2) circle[radius=2pt];
  \ifdim #3pt>#4pt
    \node[anchor=west,text=atomMark] at ({#3+0.6},#2) {#3};
    \node[anchor=east,text=compMark] at ({#4-0.6},#2) {#4};
  \else
    \node[anchor=east,text=atomMark] at ({#3-0.6},#2) {#3};
    \node[anchor=west,text=compMark] at ({#4+0.6},#2) {#4};
  \fi}
\grouprow{Claude Fable 5.2}{0}{59.89}{49.55}
\grouprow{GPT-6 Astra}{-1}{59.06}{48.15}
\grouprow{GPT-6.1 Sol}{-2}{54.03}{47.61}
\grouprow{Kimi K3.1 Pro}{-3}{50.56}{42.12}
\grouprow{GPT-6.1 Terra}{-4}{39.41}{35.83}
\grouprow{Qwen3.8-Max}{-5}{37.17}{31.94}
\grouprow{Gemini 4 Argon}{-6}{35.47}{28.29}
\grouprow{DeepSeek-V4.1-Flash}{-7}{20.24}{26.34}
\grouprow{GPT-6 Luna}{-8}{7.86}{5.11}
\end{tikzpicture}
\caption{\textbf{Atomic and composite task performance.} Filled teal points denote the mean composite score on the 25 atomic tasks and open copper points denote the mean on the 25 composite tasks. Each line connects the two means for one model.}
\label{fig:atomic-composite}
\end{figure}
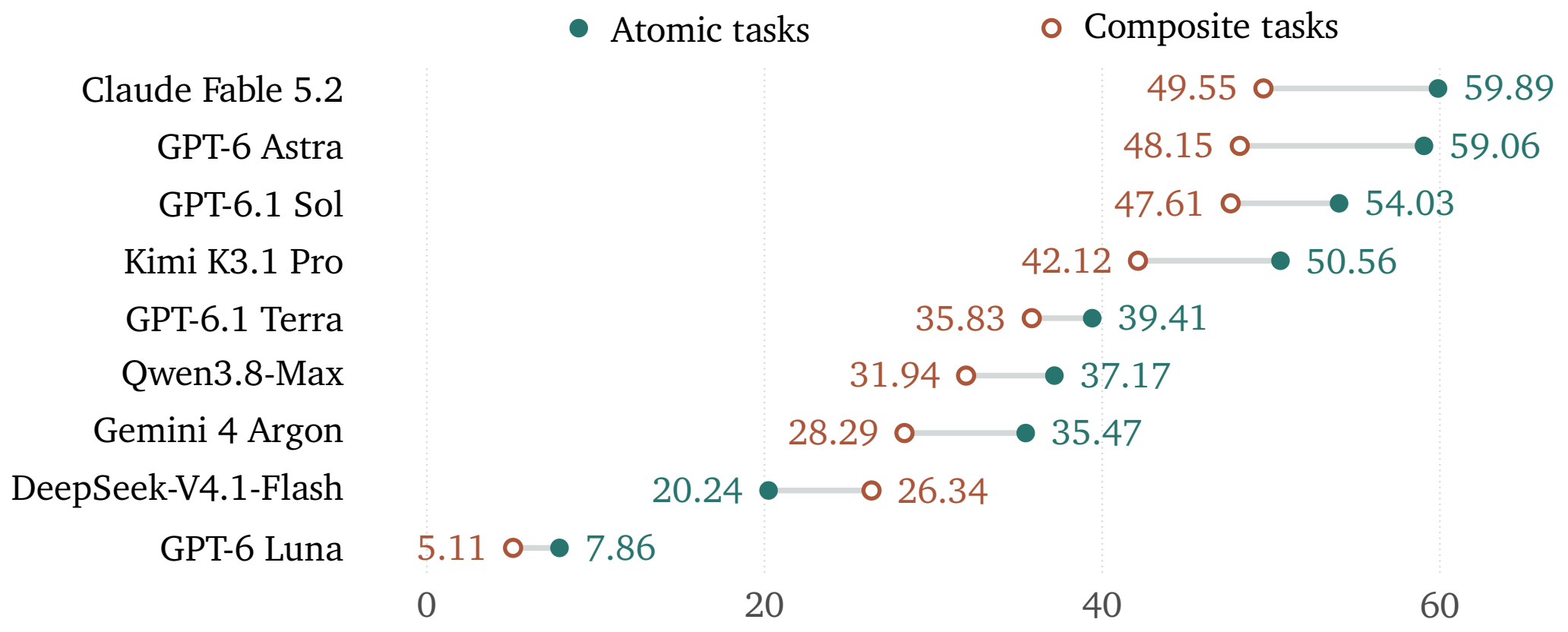

Figure~\ref{fig:atomic-composite} shows that eight of the nine models score higher on the atomic tasks. Fable leads both groups, scoring 59.89 on atomic tasks and 49.55 on composite tasks. Astra follows at 59.06 and 48.15, respectively. Sol trails Astra by only 0.54 points on composite tasks, compared with 5.04 points on atomic tasks. DeepSeek shows the reverse pattern across task types, scoring 20.24 on atomic tasks and 26.34 on composite tasks. The groups contain different objects and goals as well as different stage structures, so these differences cannot be attributed to task length alone.

\subsection{Effects of Tracking and Collision on the Composite Score}
\label{sec:stage-score}

\begin{figure}[!htb]
\centering
\begin{tikzpicture}[x=1cm,y=1cm,font=\footnotesize,inner sep=0pt]
\definecolor{retained}{HTML}{28645E}
\definecolor{trackLoss}{HTML}{945039}
\definecolor{collLoss}{HTML}{D89B6D}
\definecolor{dualLoss}{HTML}{82718C}
\definecolor{passLoss}{HTML}{AAB6B8}
\pgfmathsetmacro{\barwidth}{13.0}
\pgfmathsetmacro{\splitx}{\barwidth*37.70185125/63.8845898}
\node[anchor=west,font=\small\bfseries] at (0,2.56) {Mean task score\quad 63.88};
\node[anchor=east,font=\small\bfseries,text=trackLoss] at (\barwidth,2.56) {Reduction\quad 26.18};
\fill[retained] (0,1.78) rectangle (\splitx,2.31);
\node[text=white] at ({\splitx/2},2.045) {Composite score\quad\textbf{37.70}};
\foreach \lo/\hi/\shade/\val/\ink/\lab in {
  0.00000/5.06501/trackLoss/5.07/white/{Tracking only},
  5.06501/12.87231/collLoss/7.81/black/{Collision only},
  12.87231/22.12586/dualLoss/9.25/white/{Both},
  22.12586/26.18274/passLoss/4.06/black/{Neither}}{
  \pgfmathsetmacro{\xa}{\barwidth*\lo/26.18274}
  \pgfmathsetmacro{\xb}{\barwidth*\hi/26.18274}
  \pgfmathsetmacro{\ta}{\splitx+(\barwidth-\splitx)*\lo/26.18274}
  \pgfmathsetmacro{\tb}{\splitx+(\barwidth-\splitx)*\hi/26.18274}
  \fill[\shade!23] (\ta,1.78) .. controls (\ta,1.35) and (\xa,1.18) .. (\xa,.75)
    -- (\xb,.75) .. controls (\xb,1.18) and (\tb,1.35) .. (\tb,1.78) --cycle;
  \draw[\shade!65,line width=.32pt] (\ta,1.78) .. controls (\ta,1.35) and (\xa,1.18) .. (\xa,.75);
  \fill[\shade] (\ta,1.78) rectangle (\tb,2.31);
  \fill[\shade] (\xa,.20) rectangle (\xb,.75);
  \draw[white,line width=.6pt] (\ta,1.78)--(\ta,2.31);
  \draw[white,line width=.6pt] (\xa,.20)--(\xa,.75);
  \node[text=\ink,font=\small\bfseries] at ({(\xa+\xb)/2},.475) {\val};
  \node[anchor=north] at ({(\xa+\xb)/2},.06) {\lab};}
\draw[passLoss!65,line width=.32pt] (\barwidth,1.78)--(\barwidth,.75);
\end{tikzpicture}
\caption{\textbf{Breakdown of the 26.18-point reduction.} Labels indicate which stage checks failed. The collision check includes peak excess penetration.}
\label{fig:stage-score}
\end{figure}
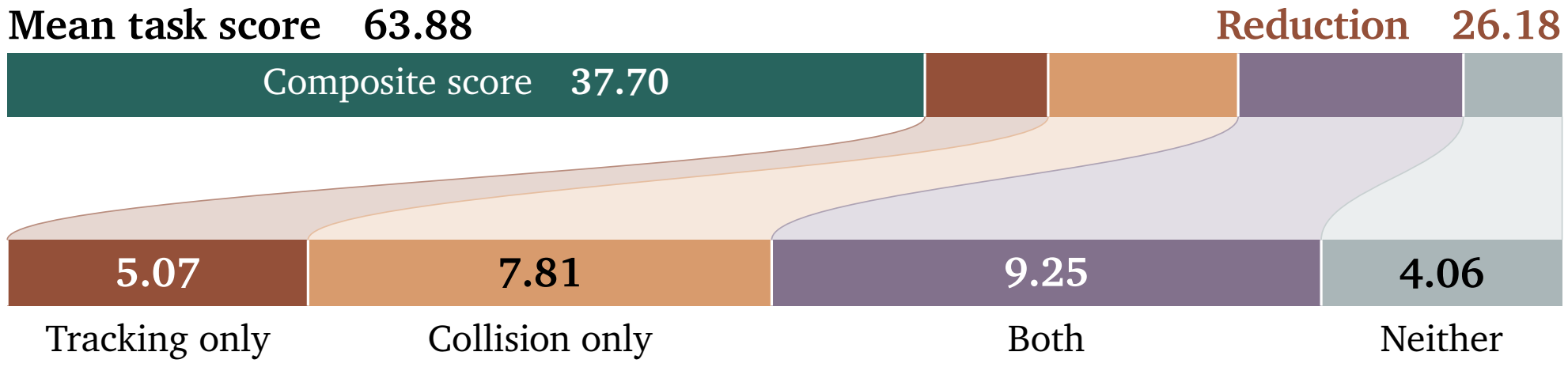

Across the 450 model--task pairs, the mean task score is 63.88. Accounting for action quality through kinematic tracking accuracy and arm--scene collision avoidance reduces the mean composite score to 37.70. Figure~\ref{fig:stage-score} breaks down the 26.18-point reduction using the same stage weights as the composite score. Of the 1,233 stages, 922 have a positive task-stage score. The other 311 contribute zero to both scores.

The 94 stages that fail only the tracking check account for 5.07 points of the reduction. On the sink-spout task, Sol specifies the required leftward turn and earns all 100 task-stage points. Its reconstructed motion incurs no collision penalty ($C_i=100$), but the arm follows the proposed poses poorly ($W_i=24.18$). Because tracking falls below 40, the stage retains only 10 points (Fig.~\ref{fig:case-spout}). The collision check captures the other aspect of action quality, arm--scene collision avoidance. In the bowl-stacking plan, for example, an arm link penetrates the cabinet shelf as the robot moves through the scene (Fig.~\ref{fig:overview}). The planned release illustrates accurate spatial goal grounding, even though the reconstructed arm collides with the shelf.

The largest reduction, 9.25 points, comes from the 165 stages that fail both robot checks. Another 7.81 points come from 125 stages that fail only the collision check. The 538 stages that pass both checks account for the remaining 4.06 points because their tracking and collision scores still scale the robot-dependent contribution.

\subsection{Alignment with Closed-Loop Execution}
\label{sec:rq4}

Having assessed the five dimensions in Table~\ref{tab:benchmark_comparison} jointly, we test whether the resulting composite score aligns with task success under a shared executor. We use the RoboHarness-Zero harness introduced by \citet{shen2026roboharnesszero} for rollouts on the same 50 RoboCasa tasks. The harness lets the VLM monitor execution and revise its instructions when progress stalls. All planners share the same frozen $\pi_{0.5}$ policy for low-level control \citep{black2025pi05}.

\begin{figure}[!htb]
\centering
\begin{tikzpicture}[x=1cm,y=1cm,font=\footnotesize]
\node[anchor=west,font=\small\bfseries] at (0,.53) {Offline composite score};
\node[anchor=west,font=\small\bfseries] at (4.95,.53) {Closed-loop success};
\node[anchor=east] at (-.14,-0.00) {GPT-6 Astra};
\fill[black!6] (0,-0.095) rectangle (3.10,0.095);
\fill[teal!65!black] (0,-0.095) rectangle (1.6618,0.095);
\node[anchor=west] at (3.16,-0.00) {53.60};
\fill[black!6] (4.95,-0.095) rectangle (8.05,0.095);
\path[fill=blue!42!black] (4.95,-0.095) rectangle (6.7480,0.095);
\node[anchor=west] at (8.11,-0.00) {58\%};
\node[anchor=east] at (-.14,-0.47) {GPT-6.1 Sol};
\fill[black!6] (0,-0.565) rectangle (3.10,-0.375);
\fill[teal!65!black] (0,-0.565) rectangle (1.5754,-0.375);
\node[anchor=west] at (3.16,-0.47) {50.82};
\fill[black!6] (4.95,-0.565) rectangle (8.05,-0.375);
\path[fill=blue!42!black] (4.95,-0.565) rectangle (6.6240,-0.375);
\node[anchor=west] at (8.11,-0.47) {54\%};
\node[anchor=east] at (-.14,-0.94) {Kimi K3.1 Pro};
\fill[black!6] (0,-1.035) rectangle (3.10,-0.845);
\fill[teal!65!black] (0,-1.035) rectangle (1.4366,-0.845);
\node[anchor=west] at (3.16,-0.94) {46.34};
\fill[black!6] (4.95,-1.035) rectangle (8.05,-0.845);
\path[fill=blue!42!black] (4.95,-1.035) rectangle (6.3140,-0.845);
\node[anchor=west] at (8.11,-0.94) {44\%};
\node[anchor=east] at (-.14,-1.41) {Qwen3.8-Max};
\fill[black!6] (0,-1.505) rectangle (3.10,-1.315);
\fill[teal!65!black] (0,-1.505) rectangle (1.0712,-1.315);
\node[anchor=west] at (3.16,-1.41) {34.56};
\fill[black!6] (4.95,-1.505) rectangle (8.05,-1.315);
\path[fill=blue!42!black] (4.95,-1.505) rectangle (6.0660,-1.315);
\node[anchor=west] at (8.11,-1.41) {36\%};
\node[anchor=east] at (-.14,-1.88) {DeepSeek-V4.1-Flash};
\fill[black!6] (0,-1.975) rectangle (3.10,-1.785);
\fill[teal!65!black] (0,-1.975) rectangle (0.7219,-1.785);
\node[anchor=west] at (3.16,-1.88) {23.29};
\fill[black!6] (4.95,-1.975) rectangle (8.05,-1.785);
\path[fill=blue!42!black] (4.95,-1.975) rectangle (5.9420,-1.785);
\node[anchor=west] at (8.11,-1.88) {32\%};
\node[anchor=east] at (-.14,-2.35) {GPT-6 Luna};
\fill[black!6] (0,-2.445) rectangle (3.10,-2.255);
\fill[teal!65!black] (0,-2.445) rectangle (0.2010,-2.255);
\node[anchor=west] at (3.16,-2.35) {6.48};
\fill[black!6] (4.95,-2.445) rectangle (8.05,-2.255);
\path[fill=blue!42!black] (4.95,-2.445) rectangle (5.8180,-2.255);
\node[anchor=west] at (8.11,-2.35) {28\%};
\end{tikzpicture}
\caption{\textbf{Offline composite scores and reported closed-loop success across 50 tasks.}}
\label{fig:matched-execution}
\end{figure}
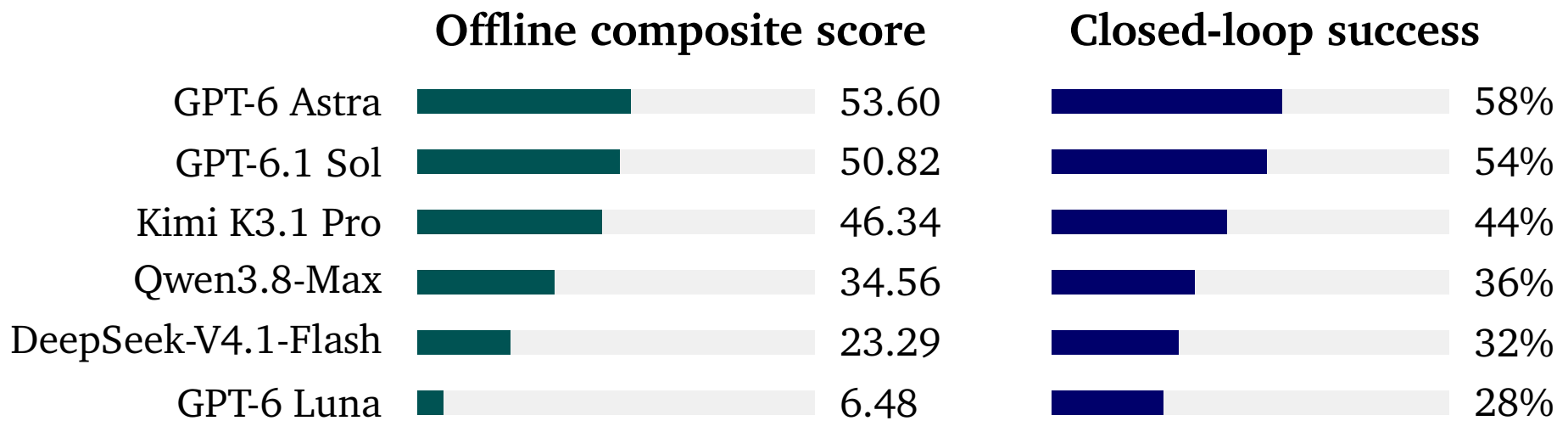

Across the six models in Figure~\ref{fig:matched-execution}, mean offline composite score and reported closed-loop success rate have identical rankings (Spearman's $\rho=1.00$) and a strong positive linear correlation (Pearson's $r=0.93$). GPT-6 Astra and GPT-6.1 Sol lead on both measures. Their offline scores differ by 2.79 points (53.60 versus 50.82), and their reported closed-loop success rates differ by four percentage points (58\% versus 54\%). Kimi K3.1 Pro follows at 44\%, ahead of Qwen3.8-Max at 36\%, DeepSeek-V4.1-Flash at 32\%, and GPT-6 Luna at 28\%. The ordering persists across the score range, even as the gaps between adjacent entries change.

\paragraph{Task-level alignment.}
The agreement in the overall ranking also appears on individual tasks. To examine this pattern, we compared pairs of models with recorded outcomes on the same task. When one model succeeded and the other failed, the successful model had the higher offline score in 62\% of the comparisons. The coffee-serving task illustrates the more common pattern. Astra and Sol both scored above 99 offline and moved the mug from the dispenser to the counter in simulation. Luna scored 1.35, and its mug remained under the dispenser (Fig.~\ref{fig:case-mug}).

\paragraph{Role of the shared executor.}
The less common reversals help explain why the gaps between models narrow in simulation. Most of these reversals occurred on tasks that $\pi_{0.5}$ could complete with the instruction alone. For example, Luna scored just 0.61 for its offline plan to move an onion from a plate to a pan, yet both Luna's closed-loop run and the standalone executor completed the transfer (Fig.~\ref{fig:case-onion}). Across the suite, the executor alone solved 21 of 50 tasks, including 13 of Luna's 14 successes. Its manipulation skills can therefore support a planner even when the trajectory proposed offline is poor.

Execution can also break down after a strong offline plan. Astra scored 99.61 for placing a piece of cake in a mixer bowl, but during the closed-loop run the cake slipped from the gripper and landed beside the bowl (Fig.~\ref{fig:case-cake}). Together, these cases illustrate a limitation of closed-loop evaluation. The success rate combines the quality of the planner's decisions with the downstream policy's ability to carry them out.

\begin{figure}[!htb]
\centering
\input{figures/paired-cost}
\caption{Mean cost per task for GPT-6.1 Sol on ten atomic and ten composite tasks. Uncached input and generated output tokens use the left axis. API waiting time and total runtime share the right axis.}
\label{fig:paired-cost}
\end{figure}
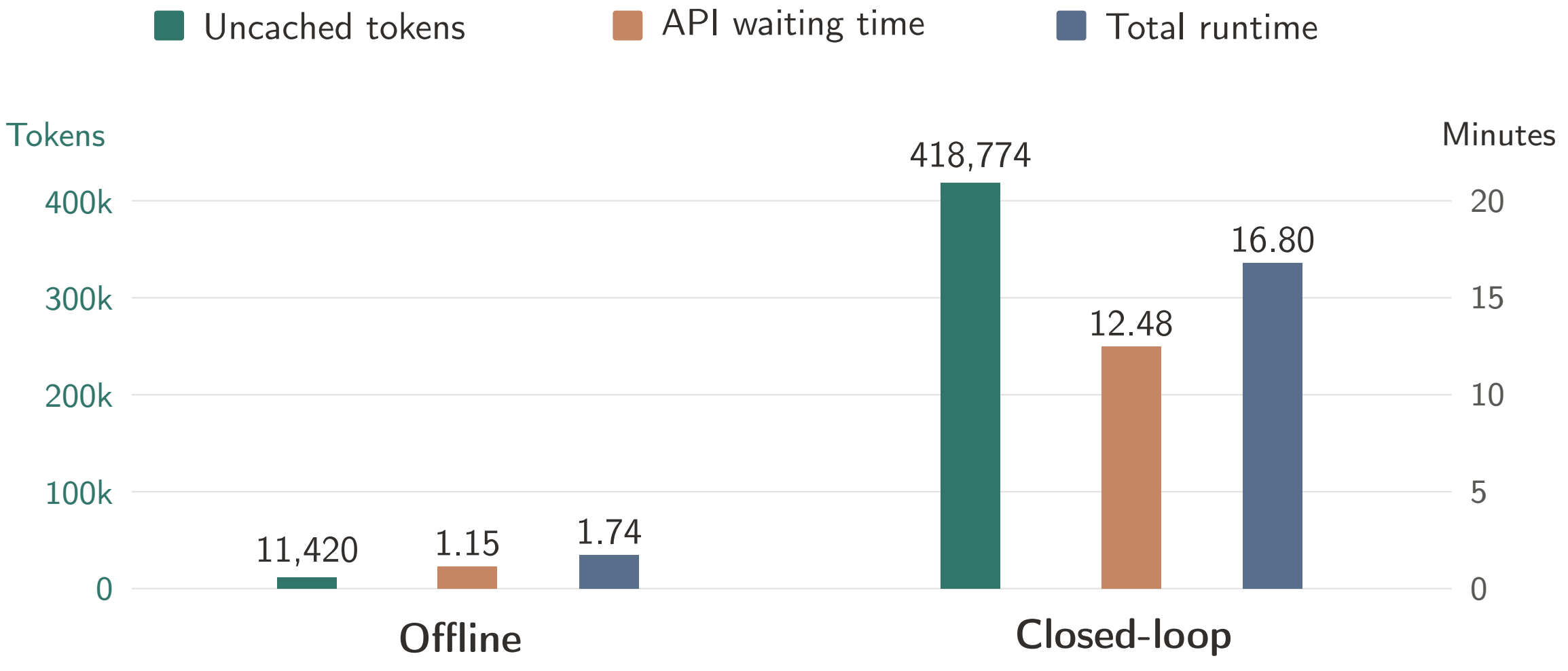

\paragraph{Model requests and token use.}
VRBench requests a complete trajectory from the initial observations, while the closed-loop harness queries the VLM as new observations arrive. Including retries, the two settings averaged 1.05 and 85.15 API requests per task, respectively, on the paired sample (Figure~\ref{fig:paired-cost}).
Comparing GPT-6.1 Sol at medium reasoning effort on the same ten atomic and ten composite tasks, total input and output tokens averaged 11,806 per task offline versus 14.18 million in closed loop (97.0\% cached inputs).
Counting only uncached input and generated output, VRBench reduced token use by 97.3\%, from 418,774 to 11,420 tokens per task ($37\times$ fewer).
Summed across requests within each task, including retries, API waiting time averaged 1.15 minutes offline versus 12.48 minutes in closed loop—a 90.8\% reduction ($11\times$ faster)—with an even larger gap on composite tasks (1.28 versus 17.56 minutes) than on atomic tasks (1.02 versus 7.41 minutes).

\paragraph{Runtime on the evaluation hardware.}
We compared per-task runtimes on the same MacBook Air (Apple M5, 10-core CPU, 10-core GPU, 32\,GB unified memory). For each offline task, IK queries were batched on the GPU, while graph search, tracking checks, and collision scoring ran on the CPU. Model requests averaged 69.09 seconds, joint reconstruction 30.60 seconds, and scoring 4.78 seconds, for a total of 1.74 minutes per task.
Closed-loop episodes ran one at a time, with physics simulation on the CPU and the $\pi_{0.5}$ executor on the GPU. They averaged 16.80 minutes per task, or $9.65\times$ the offline time. Offline and closed-loop times averaged 1.72 and 8.93 minutes on atomic tasks, and 1.76 and 24.66 minutes on composite tasks, respectively. These are per-task timings on this hardware. Faster hardware can shorten local computation in both settings, and additional devices can reduce batch runtime by processing tasks in parallel.

\section{Limitations}
\label{sec:limitations}
VRBench scores a plan from the information available before motion begins. The initial views can leave a handle occluded or the depth of a shelf ambiguous, and the model has no opportunity to resolve that ambiguity by moving the camera. Reconstructing the arm trajectory lets us test pose tracking and sample for collisions, but a feasible approach to an object is only one condition for a successful grasp. Contact forces, slippage during transport, and recovery from a failed grasp require observations during execution. Conversely, a capable downstream policy can complete a task for which the VLM proposed a poor initial trajectory. A composite score therefore measures the quality of the proposed plan, and its numerical value should not be interpreted as a probability of task success. This distinction also matters when comparing scores with closed-loop outcomes, since the downstream policy can influence how closely task success aligns with offline planning scores.

\section{Conclusion}
VirtualRoboBench (VRBench) evaluates VLM robotic manipulation capabilities offline by asking a model to express an entire task as a 6-DoF end-effector trajectory with gripper and contact events.
Evaluating stage goals and ordering alongside inverse-kinematics tracking and arm--scene collision risk captures five coupled manipulation dimensions in a single rollout-free pass.
Across nine frontier VLMs, incorporating robot motion quality exposes major feasibility failures missed by task-level checks alone—separating models with highest semantic task progress (GPT-6 Astra) from those producing the most robot-compatible plans overall (Claude Fable 5.2).

Comparison with closed-loop simulation under a shared executor shows that VRBench scores strongly correlate with actual task success rates (Spearman's $\rho=1.00$, Pearson's $r=0.93$) and predict the better planner on most decisive individual tasks. In the paired cost study, offline evaluation used 97\% fewer uncached tokens and took about one-tenth of the closed-loop time per task on our hardware.
VRBench thus provides a practical, execution-aligned benchmark for rapidly assessing VLM manipulation capabilities without simulator rollouts.

\medskip
\renewcommand{\bibfont}{\small}
\bibliographystyle{plainnat}
\bibliography{references}

\clearpage\appendix

\section{Joint Reconstruction}
\label{app:joint-reconstruction}

\paragraph{Inputs and outputs.}
The VLM specifies the pose of the tool, but several arm configurations can place the tool at the same pose. Its choice at one waypoint can determine whether the arm can reach the next. Given the robot kinematics, joint limits $\mathcal Q$, initial configuration $q^0$, and the predicted TCP poses, we seek a single joint path through the ordered targets:
\begin{equation}
q_0=q^0,\qquad q_j\in\mathcal Q,\qquad \FK(q_j)\approx\hat T_j .
\label{eq:IK}
\end{equation}
Here $\hat T_j$ is a reference pose and $\FK$ maps the arm configuration to its TCP pose. The output includes the joint path, the index of each original waypoint on that path, and the outcome of the tracking check. The indices preserve the alignment with the VLM's gripper and contact events for stage scoring.

\paragraph{Choice of method.}
Prior work addresses different aspects of this ambiguity. IKFlow generates multiple configurations for a single pose using a learned distribution \citep{ames2022ikflow}. CppFlow uses generative IK to initialize path search and refinement \citep{cppflow2024}, while \citet{yoon2023initialization} learn initial joint trajectories for a path-following optimizer. TORM optimizes tracking, smoothness, and obstacle avoidance jointly \citep{kang2020torm}, and cuRobo uses parallel optimization to generate collision-free motions \citep{sundaralingam2023curobo}. For VRBench, collision risk is itself a measurement: allowing the solver to reroute the tool around an obstacle could hide a collision in the predicted motion. We therefore fix the Cartesian reference and evaluate collisions on the reconstructed arm afterward. We adopt the layered search of IKLink \citep{iklink2024} and the guided sampling strategy of \citet{wang2025anytime}. This combination considers alternative arm configurations along the supplied path and uses numerical IK without requiring a trained solver for the robot.

\paragraph{Guided search on the RoboCasa suite.}
We prepend the initial TCP pose $\FK(q^0)$ to the 36 predicted waypoints. We subdivide each segment by linear position interpolation and spherical orientation interpolation, using increments of at most 10\,mm and 0.05\,rad. Each reference pose defines a graph layer. We seed sparse anchor layers with damped least-squares IK solutions, use a coarse path to guide sampling in intervening layers, and add random seeds to explore other configurations. Dynamic programming links candidates across layers, first minimizing transitions that exceed the joint-velocity allowance under the solver's virtual timing, then minimizing accumulated joint travel on the other transitions. The first layer is fixed at $q^0$. We accept the first complete path that respects joint bounds and tracks the reference within 2\,mm and 0.03\,rad at the nodes and at sampled points along joint interpolation. The GPU implementation uses fixed search budgets of 96 rounds and 20,000 planner-requested IK calls per trajectory, with no wall-clock cutoff.

\paragraph{Fallback and interpretation.}
If the guided search yields no path passing these checks, the fallback works directly on the initial pose and the 36 original waypoints. It samples several IK candidates per waypoint, progressively relaxes pose tolerances, and targets gaps between adjacent candidate sets, with a separate budget of 10,000 IK calls. It first seeks a sequence of accepted candidates with changes of at most 0.5\,rad per joint between waypoints. If needed, it admits approximate candidates and finally relaxes the step bound. A complete, finite fallback path can therefore be scored even when strict tracking fails. We record which path was selected and measure its actual tracking error and joint changes.

\paragraph{Changing the task suite.}
Joint reconstruction is specific to the robot and task suite. Another suite can supply its kinematics and TCP frame, select suitable tolerances and search budgets, or substitute a solver suited to its embodiment. The evaluator needs a joint path anchored at the supplied initial state, its correspondence to the predicted waypoints, and its reconstruction status. Within each suite, the same solver and settings apply to every model.

\section{Scoring Details}
\subsection{Task Scoring}
\label{app:task-scoring}

\paragraph{Distance and scale.}
Each stage specifies what to measure and when. For a point-to-region check, let $E_i$ contain the eligible samples and $x_k$ be the point tested at sample $k$. Then $d_i=\min_{k\in E_i}\operatorname{dist}(x_k,\Omega_i)$, where the distance is zero inside the accepted volume. Multiple allowed regions form a union. The tested geometry depends on the interaction: a grasp can use the separation between the finger pads and the target, while placement can use a point on the held object transformed to its predicted release pose. For explicit distance checks, we choose $\sigma_i$ for each stage with the size of the manipulated object or target part in mind. For example, the kettle task uses $10$\,mm for the switch and $50$\,mm for the lid. The scale controls how quickly the score falls outside the accepted region. Orientation enters only where the operation requires it. For example, when releasing the blender lid, let $g$ be the distance of its tracked point from the target and $\phi$ its tilt relative to the grasp pose. The residual is $\max\{g,\,(0.025\,\mathrm{m/rad})[\phi-\pi/6]_+\}$, with $g$ in meters, $\phi$ in radians and $[z]_+=\max(z,0)$. This stage uses $\sigma_i=50$\,mm.

\paragraph{Completion and aggregation.}
For the suite's explicit distance checks, $h_i=1$ when an eligible sample has zero residual, up to $10^{-9}$\,m for numerical comparisons. The accepted region and any specified contact margin already enter the residual. Thus a release 1\,mm outside its accepted region with $\sigma_i=50$\,mm earns $100e^{-1/50}\approx98.02$ points if all other conditions hold, but leaves a dependent stage blocked. Operations spanning several samples can specify their own score and completion condition. For example, after the sponge is grasped, wiping requires five distinct contact locations at least 20\,mm apart and a 100\,mm diagonal of their planar bounding box. With $n$ retained locations and diagonal $L$ in millimeters, the stage earns $100\min(1,n/5,L/100)$ and completes when both targets are met. Each stage has unit weight, including a blocked stage, so a task with $m$ stages assigns each a share $1/m$. Where the instruction permits alternative assignments, the evaluator scores each allowed assignment and takes the highest task score. The dataset supplies the regions, event requirements, tolerances and dependencies for each task in \texttt{grading/regions.json}.

\subsection{Robot Scoring}
\label{app:robot-scoring}

This section gives the tracking and collision scores for the joint path selected in Appendix~\ref{app:joint-reconstruction}.

For reconstructed TCP pose $(p_j,Q_j)$ and predicted pose $(\hat p_j,\hat Q_j)$, we calculate position and orientation errors as
\begin{align}
\epsilon_j^p
&=\|p_j-\hat p_j\|_2,\nonumber\\
\epsilon_j^r
&=\arccos\!\left[
\clip\!\left(
\frac{\operatorname{tr}(Q_j^\top\hat Q_j)-1}{2},
-1,1
\right)\right].
\label{eq:tracking}
\end{align}

\paragraph{Kinematic Tracking.}
Tracking is evaluated at the $K$ waypoints predicted by the VLM. Let $n_k$ denote the reconstruction node corresponding to predicted waypoint $k$, with $n_0=0$ representing the initial robot configuration.

At each predicted waypoint, we compute the errors in Eq.~\eqref{eq:tracking} using $\FK(q_{n_k})$. We also measure the largest single-joint change from the preceding waypoint:
\begin{equation}
u_k=\|q_{n_k}-q_{n_{k-1}}\|_\infty.
\label{eq:joint-change}
\end{equation}

For an evaluation interval $I$, let $e_p(I)$ and $e_r(I)$ be the root-mean-square position and orientation errors over its assigned waypoints, measured in millimeters and degrees. Let $u_{95}(I)$ and $u_{\max}(I)$ denote the 95th percentile and maximum joint changes, measured in degrees.

Omitting $I$ on the right-hand side for readability, we define
\begin{align}
P_I &= \exp(-e_p/\lambda_p),\nonumber\\
O_I &= \exp(-e_r/\lambda_r),\nonumber\\
J_I &= \exp\!\left[
-\beta_J\left(
\frac{[u_{95}-\tau_{95}]_+}{\tau_{95}}
+\frac{[u_{\max}-\tau_{\max}]_+}{\tau_{\max}}
\right)\right],\nonumber\\
W_I &= 100J_I\sqrt{P_IO_I},
\label{eq:kinematic-weight}
\end{align}
where $[x]_+=\max(x,0)$.

The parameters $\lambda_p$ and $\lambda_r$ control the penalties for position and orientation errors. The parameters $\tau_{95}$, $\tau_{\max}$, and $\beta_J$ determine how large joint changes affect the score.

The resulting tracking score satisfies $W_I\in[0,100]$. Higher scores indicate better tracking of the predicted TCP poses with fewer large joint changes.

Because models are prompted to specify the full motion trajectory across $K$ ordered waypoints rather than only sparse endpoints, large single-step joint jumps ($u_k > \tau_{95}$) indicate underspecified motion segments or sudden configuration flips between adjacent poses and are penalized through $J_I$.
When a trajectory exceeds the strict tracking tolerance of guided search and triggers the fallback solver (Appendix~\ref{app:joint-reconstruction}), $P_I$, $O_I$, and $J_I$ smoothly quantify the resulting pose deviation and joint discontinuity rather than discarding the prediction.

\paragraph{Collision Checking.}
We replay the reconstructed robot arm against the scene geometry at the corresponding expert reference frames. Each joint-space segment is subdivided so that the maximum single-joint change between consecutive collision samples does not exceed $\Delta_{\mathrm{col}}$.

Collision checking considers eligible arm--environment pairs, with task-specific intended contacts excluded. Gripper geometry and robot self-collision do not contribute to the current collision score.

Let $\delta_{b\ell j}$ denote the signed separation, in millimeters, between robot link $b$ and obstacle $\ell$ at collision sample $j$. Negative values indicate penetration.

With geometric tolerance $m_{\mathrm{col}}$, the excess penetration at each sample is
\begin{equation}
p_j=\max\!\left(
\{0\}\cup
\{-\delta_{b\ell j}-m_{\mathrm{col}}:
(b,\ell)\text{ is eligible at }j\}
\right).
\label{eq:penetration}
\end{equation}

For an evaluation interval $I$ containing $N_I$ collision samples, we calculate
\begin{align}
\bar p_I
&=\frac{1}{N_I}\sum_{j\in I}p_j,\nonumber\\
p_I^{\max}
&=\max_{j\in I}p_j,\nonumber\\
f_I
&=\frac{1}{N_I}
\sum_{j\in I}\mathbf{1}[p_j>0].
\label{eq:collision-statistics}
\end{align}

Here, $\bar p_I$ is the average excess penetration, $p_I^{\max}$ is the largest excess penetration, and $f_I$ is the fraction of samples exceeding the geometric tolerance.

The raw collision score for interval $I$ is
\begin{equation}
C_I=\frac{100}{3}\left[
\exp(-\bar p_I/\lambda_c)
+\exp(-p_I^{\max}/\lambda_c)
+1-f_I
\right],
\label{eq:collision-score}
\end{equation}
where $\lambda_c$ controls the penalty for excess penetration.

All collision samples, including those with zero excess penetration, contribute to the average. The resulting score satisfies $C_I\in[0,100]$, with higher values indicating lower collision risk under the evaluated geometry.

\paragraph{Stage-wise Robot Scores.}
We compute the tracking and collision scores for both the complete trajectory and individual task stages.

For each task stage $i$, let $I_i$ denote the motion interval assigned according to the task-stage requirements and their prerequisites. The motion interval of the last stage with a positive task-stage score extends to the end of the predicted trajectory so that subsequent motion is also evaluated.

The stage-level scores are
\begin{equation}
C_i=C_{I_i},\qquad
W_i=W_{I_i}.
\label{eq:CW}
\end{equation}

The composite score in Eq.~\eqref{eq:overall-score} uses the stage factors defined in Section~\ref{sec:composite}. With $m_{\mathrm{col}}=10$\,mm and $\theta_p=50$\,mm, a sampled penetration of 60\,mm for an eligible pair gives $p_j=50$\,mm and sets $\bar C_i=0$ for the stage containing that sample. The reported collision score $C$ still follows Eq.~\eqref{eq:collision-score} over the complete trajectory.

Section~\ref{sec:setup} lists the shared scoring parameters. The task files specify the stage geometry, operation conditions and collision exclusions. Because the evaluation relies on sampled geometry and finite-budget joint reconstruction, these metrics do not certify that a predicted trajectory is physically executable.

\input{appendix-cases}

\end{document}

%% file: figures/composite-score.tex
\begingroup
\definecolor{vrClay}{HTML}{B76148}
\definecolor{vrSand}{HTML}{C48269}
\definecolor{vrStone}{HTML}{BDB6A9}
\definecolor{vrGold}{HTML}{A97C24}
\definecolor{vrSilver}{HTML}{77818D}
\definecolor{vrBronze}{HTML}{A65F32}
\definecolor{vrInk}{HTML}{282522}
\noindent\makebox[\linewidth][c]{%
\resizebox{0.995\linewidth}{!}{%
\begin{tikzpicture}[x=1pt,y=.87pt,inner sep=0pt,outer sep=0pt,text=vrInk]
\definecolor{vrRuntime}{HTML}{42689C}
\definecolor{vrCost}{HTML}{A96732}
\path[use as bounding box] (0,4) rectangle (525,139);
\draw[vrStone!55,line width=.5pt] (358,8) -- (358,127);
\newcommand{\scorerow}[5]{%
  \node[anchor=east,font=\sffamily\fontsize{11.2}{12}\selectfont]
    at (113,#4) {#1};
  \pgfmathsetmacro{\barend}{120+3.20*#3}
  \fill[#5,rounded corners=1.3pt] (120,{#4-6.9}) rectangle (\barend,{#4+6.9});
  \node[anchor=west] at ({\barend+3},#4)
    {\includegraphics[page=#2,width=11pt]{figures/model-marks.pdf}};
  \node[anchor=east,text=#5,font=\rmfamily\bfseries\fontsize{13}{14}\selectfont]
    at (350,#4) {#3};}
\scorerow{\textcolor{vrGold}{\faTrophy}\enspace Claude Fable 5.2}{1}{54.72}{120}{vrClay}
\scorerow{\textcolor{vrSilver}{\faMedal}\enspace GPT-6 Astra}{2}{53.60}{99}{vrSand}
\scorerow{\textcolor{vrBronze}{\faMedal}\enspace Kimi K3.1 Pro}{3}{46.34}{78}{vrStone}
\scorerow{Qwen3.8-Max}{5}{34.56}{57}{vrStone}
\scorerow{Gemini 4 Argon}{4}{31.88}{36}{vrStone}
\scorerow{DeepSeek-V4.1-Flash}{6}{23.29}{15}{vrStone}
\foreach \tick in {0,...,4}{
  \pgfmathtruncatemacro{\minutes}{5*\tick}
  \pgfmathsetmacro{\level}{28+24*\tick}
  \draw[vrStone!45,line width=.4pt] (389,\level) -- (500,\level);
  \node[anchor=east,text=vrRuntime,font=\sffamily\scriptsize] at (385,\level) {\minutes};
  \node[anchor=west,text=vrCost,font=\sffamily\scriptsize] at (504,\level) {\tick};
}
\node[rotate=90,text=vrRuntime,font=\sffamily\bfseries\footnotesize] at (368,76) {Runtime (min)};
\node[rotate=-90,text=vrCost,font=\sffamily\bfseries\footnotesize] at (520,76) {API cost (USD)};
\newcommand{\efficiencybar}[4]{%
  \fill[#4] ({#1-6},28) rectangle ({#1+6},{28+#2});
  \node[anchor=south,text=#4,font=\sffamily\bfseries\scriptsize,fill=white,inner xsep=.6pt]
    at (#1,{31+#2}) {#3};}
\efficiencybar{404}{24*1.74/5}{1.74}{vrRuntime}
\efficiencybar{427}{24*.056316}{0.056}{vrCost}
\efficiencybar{462}{24*16.80/5}{16.80}{vrRuntime}
\efficiencybar{485}{24*3.02421}{3.02}{vrCost}
\node[font=\sffamily\bfseries\footnotesize] at (415.5,15) {Offline};
\node[font=\sffamily\bfseries\footnotesize] at (473.5,15) {Closed-loop};
\end{tikzpicture}}}\par
\endgroup

%% file: figures/paired-cost.tex
\begingroup
\definecolor{costTokens}{HTML}{31756B}
\definecolor{costAPI}{HTML}{C48663}
\definecolor{costTotal}{HTML}{596E8A}
\definecolor{costInk}{HTML}{312E2B}
\resizebox{.91\linewidth}{!}{%
\begin{tikzpicture}[x=1pt,y=1pt,inner sep=0pt,outer sep=0pt,text=costInk,font=\sffamily\small]
\foreach \x/\col/\name in {42/costTokens/Uncached tokens,165/costAPI/API waiting time,284/costTotal/Total runtime}{
  \fill[\col,rounded corners=.8pt] (\x,165) rectangle ({\x+8},173);
  \node[anchor=west] at ({\x+13},169) {\name};
}
\node[anchor=west,text=costTokens,font=\sffamily\footnotesize] at (2,140) {Tokens};
\node[anchor=east,font=\sffamily\footnotesize] at (418,140) {Minutes};
\foreach \tick/\tokenlabel/\minute in {0/0/0,1/100k/5,2/200k/10,3/300k/15,4/400k/20}{
  \pgfmathsetmacro{\tickheight}{18+26*\tick}
  \draw[costInk!14,line width=.4pt] (36,\tickheight) -- (390,\tickheight);
  \node[anchor=east,text=costTokens,font=\sffamily\footnotesize] at (31,\tickheight) {\tokenlabel};
  \node[anchor=west,text=costInk!80,font=\sffamily\footnotesize] at (395,\tickheight) {\minute};
}
\newcommand{\metricbar}[4]{%
  \fill[#4] ({#1-8},18) rectangle ({#1+8},{18+#2});
  \node[anchor=south,font=\sffamily\small,fill=white,inner xsep=1pt] at (#1,{21+#2}) {#3};}
\metricbar{83}{26*.11420}{11,420}{costTokens}
\metricbar{126}{26*1.15/5}{1.15}{costAPI}
\metricbar{164}{26*1.74/5}{1.74}{costTotal}
\metricbar{261}{26*4.18774}{418,774}{costTokens}
\metricbar{304}{26*12.48/5}{12.48}{costAPI}
\metricbar{342}{26*16.80/5}{16.80}{costTotal}
\node[font=\sffamily\bfseries] at (124,5) {Offline};
\node[font=\sffamily\bfseries] at (302,5) {Closed-loop};
\end{tikzpicture}}
\endgroup

%% file: appendix-cases.tex
\section{Case Studies}
\label{app:cases}
\newcommand{\caseimage}[3]{\begin{minipage}[t]{#1\linewidth}\centering
  #2\\[-1pt]
  {\footnotesize\sffamily #3}\end{minipage}}
\newenvironment{casepanel}{\par\smallskip\noindent\begin{minipage}{\linewidth}}{\end{minipage}\par}
\newcommand{\caseinputs}[4]{\noindent\begin{minipage}{\linewidth}
  \textbf{Instruction.} \emph{#1}\par\smallskip
  \centering #2\hspace{.018\linewidth}%
  #3\hspace{.018\linewidth}%
  #4\end{minipage}\par}
\newcommand{\caseheading}[1]{\noindent{\sffamily\bfseries #1}\par\smallskip}
\definecolor{caseTeal}{HTML}{087D84}
\definecolor{casePurple}{HTML}{AD4084}
\definecolor{caseAmber}{HTML}{AD6C09}
\newcommand{\casetraces}{{\color{caseTeal}Predicted TCP}\quad {\color{casePurple}Reconstructed TCP}\quad {\color{caseAmber}Target region}}
\subsection{Turning a sink spout}
\caseinputs{Turn the sink spout to the left.}%
{\caseimage{.225}{\includegraphics[width=\linewidth]{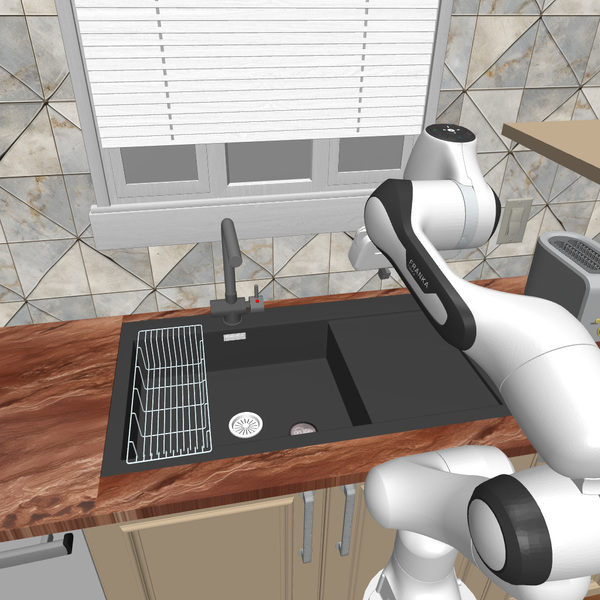}}{Left view}}%
{\caseimage{.225}{\includegraphics[width=\linewidth]{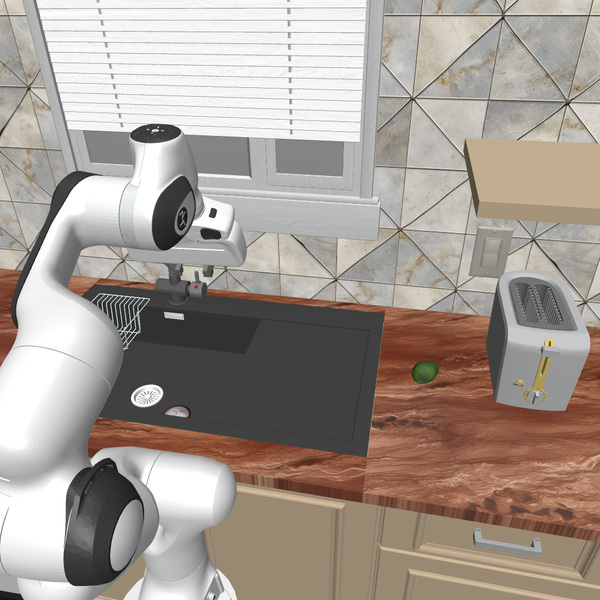}}{Right view}}%
{\caseimage{.225}{\includegraphics[width=\linewidth]{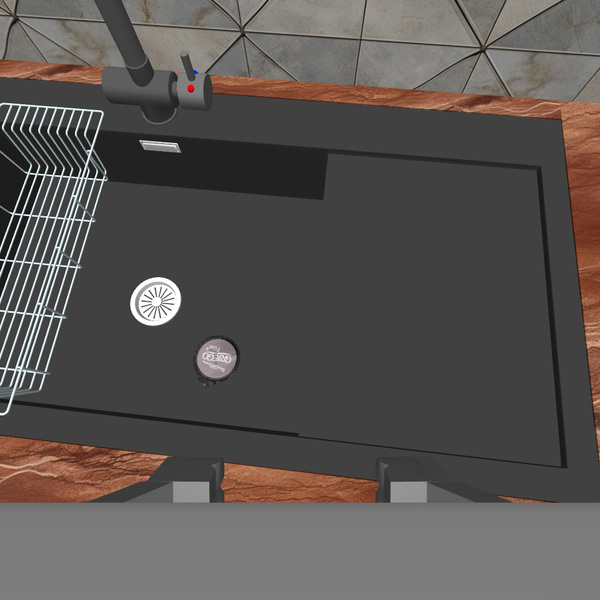}}{Wrist view}}
\begin{casepanel}
\caseheading{Predicted and reconstructed motion}
\noindent\caseimage{.325}{\includegraphics[width=\linewidth]{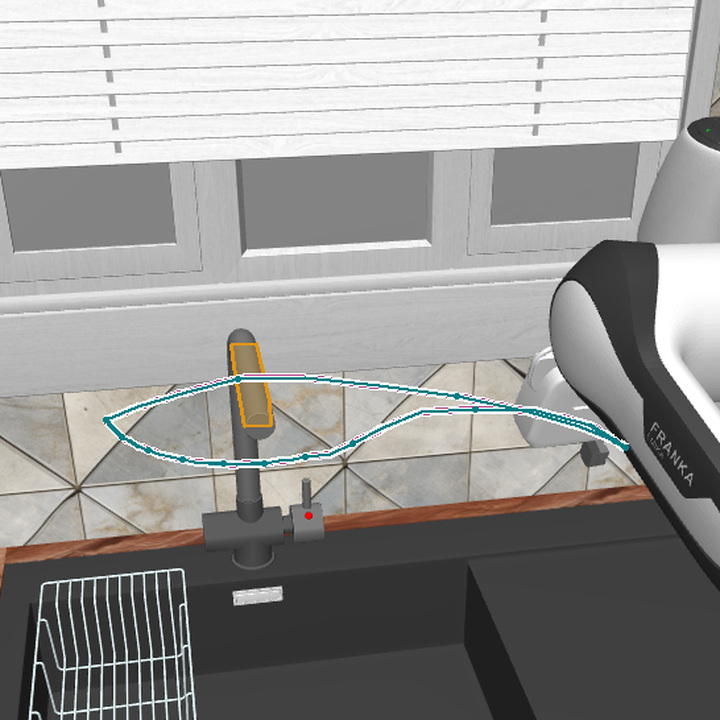}}{Fable}\hfill
\caseimage{.325}{\includegraphics[width=\linewidth]{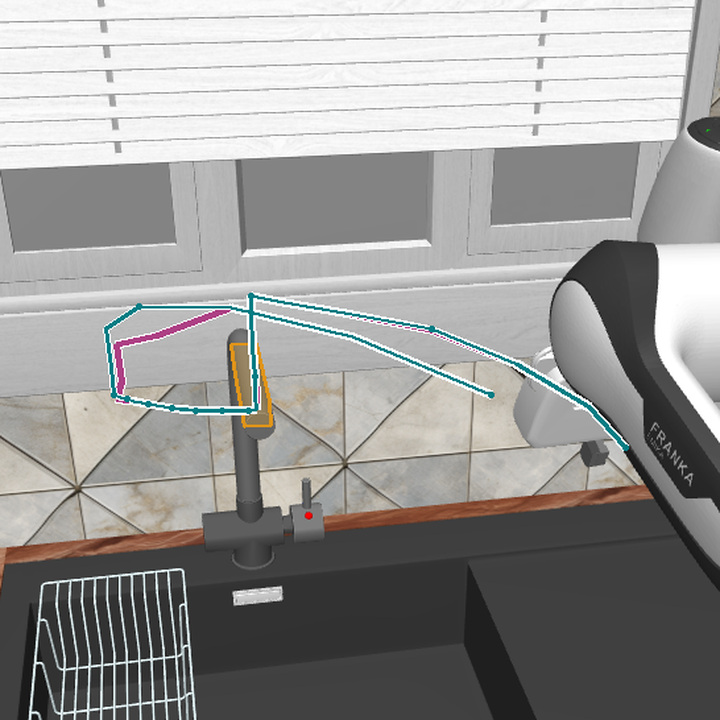}}{Astra}\hfill\caseimage{.325}{\includegraphics[width=\linewidth]{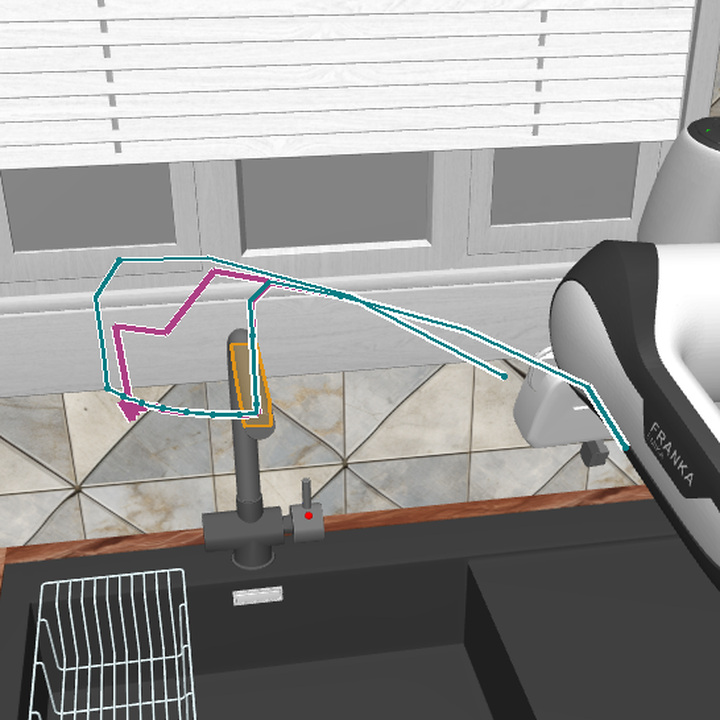}}{Sol}\par
\smallskip\centerline{\footnotesize\sffamily\casetraces}
\captionof{figure}{\textbf{The same turn can have different tracking costs.}
Fable's two curves nearly coincide: its largest 3D position error is 0.2\,mm, compared with 31.9\,mm for Astra and 85.6\,mm for Sol. Although Astra earns the higher task score (100 versus 90.48), Fable tracks the proposed turn more closely (99.39 versus 59.35) and earns the higher composite score (86.52 versus 63.42). Sol also earns 100 on the task, but its tracking score of 24.18 falls below the cutoff of 40, leaving a composite score of 10.}
\label{fig:case-spout}
\end{casepanel}

\subsection{Grasping a spatula before pressing chicken}
\caseinputs{Pick up the spatula from the plate and lightly press it against the chicken for a few seconds, then release.}%
{\caseimage{.225}{\includegraphics[width=\linewidth]{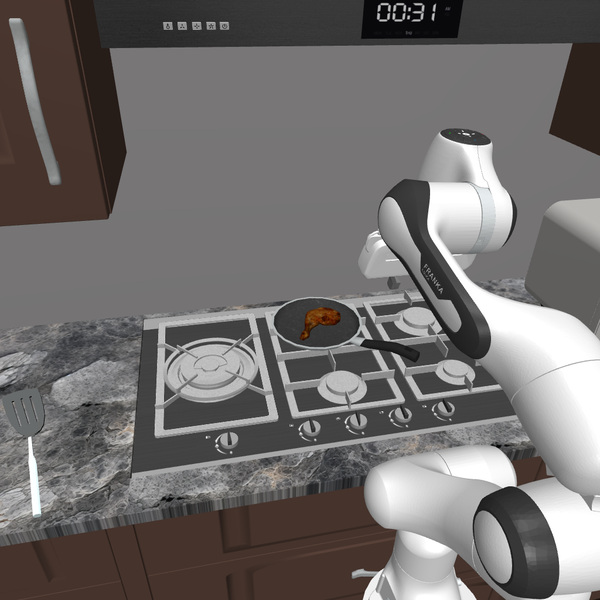}}{Left view}}%
{\caseimage{.225}{\includegraphics[width=\linewidth]{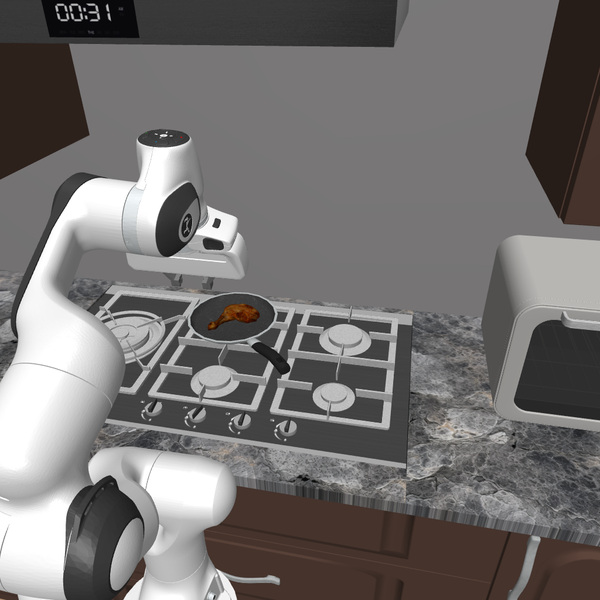}}{Right view}}%
{\caseimage{.225}{\includegraphics[width=\linewidth]{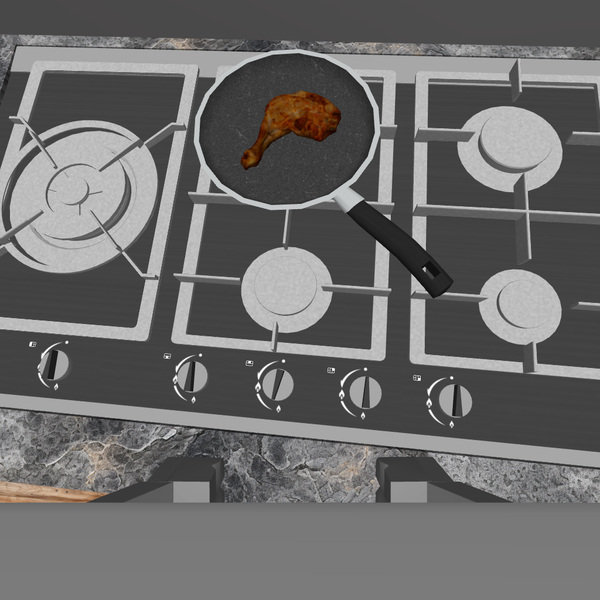}}{Wrist view}}
\begin{casepanel}
\caseheading{Argon's predicted grasp}
\noindent\hfill\caseimage{.36}{\includegraphics[width=\linewidth]{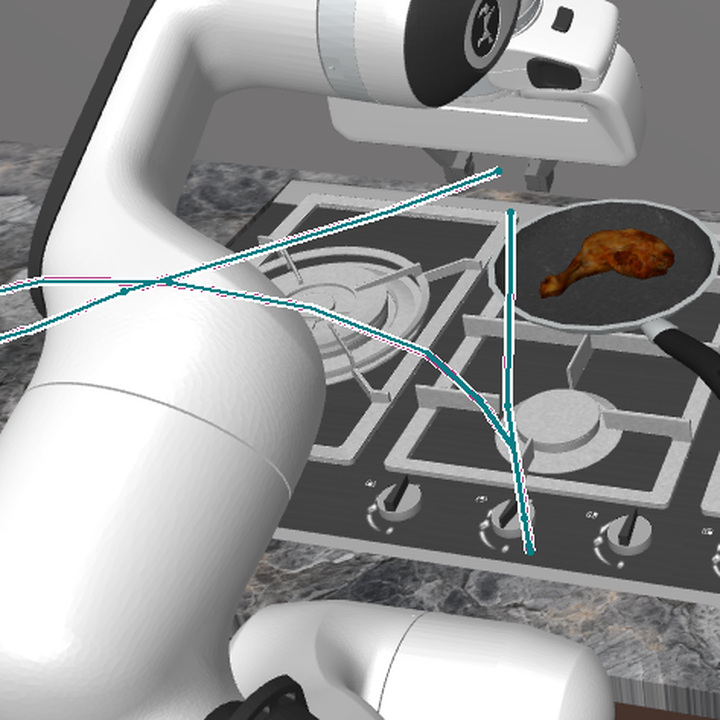}}{Planned motion, right view}\hfill
\caseimage{.36}{\includegraphics[width=\linewidth]{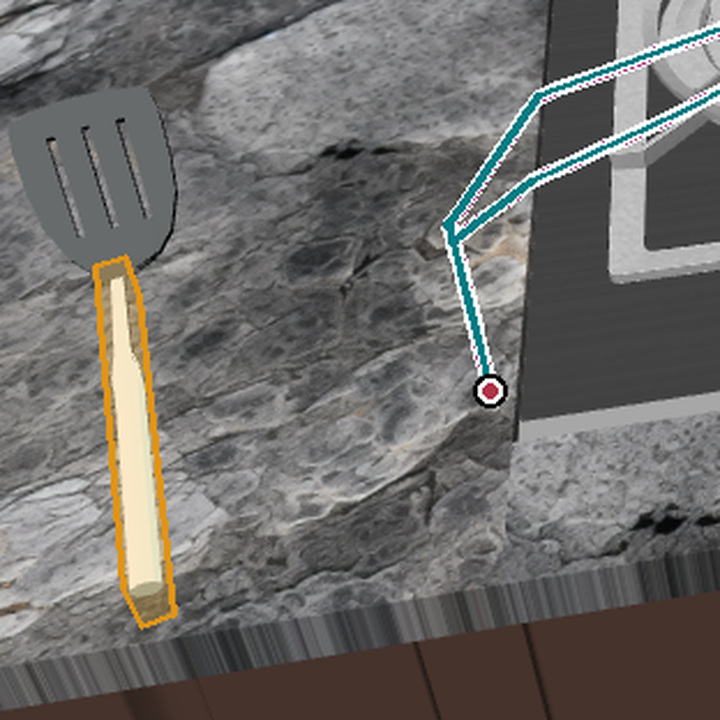}}{Handle and grasp point, left view}\hfill\par
\smallskip\centerline{\footnotesize\sffamily\casetraces}
\captionof{figure}{\textbf{Close tracking of a misplaced grasp.} The ring marks Argon's proposed grasp, 14.9\,cm from the shaded handle region in 3D. The grasp stage earns 5.05 points. Pressing requires a successful grasp, so the pressing stage earns zero. The two stages carry equal weight, giving a task score of 2.53 even though the arm tracks the proposed motion closely ($W=99.42$).}
\label{fig:case-spatula}
\end{casepanel}

\subsection{Moving a coffee mug to the counter}
\caseinputs{Pick the mug from under the coffee machine dispenser and place it on the counter.}%
{\caseimage{.225}{\includegraphics[width=\linewidth]{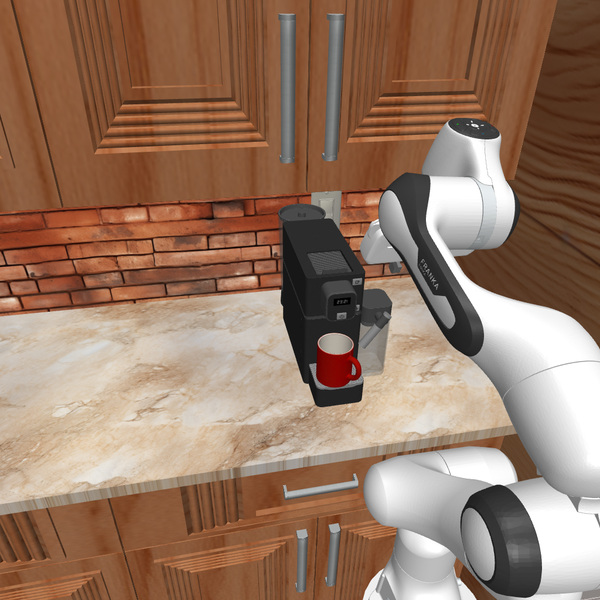}}{Left view}}%
{\caseimage{.225}{\includegraphics[width=\linewidth]{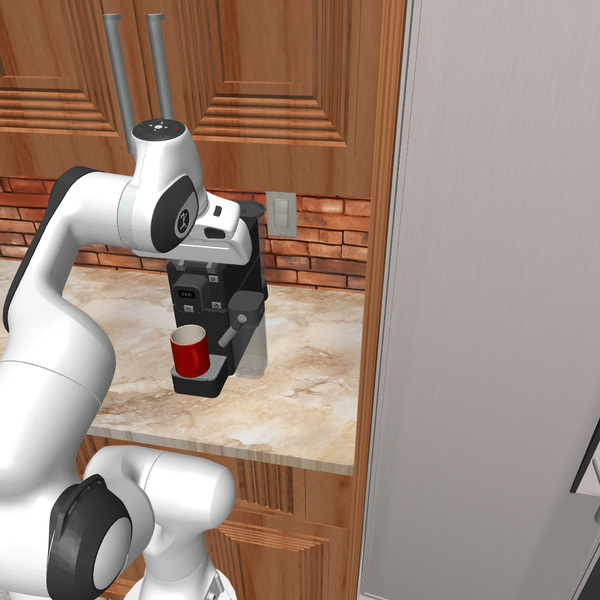}}{Right view}}%
{\caseimage{.225}{\includegraphics[width=\linewidth]{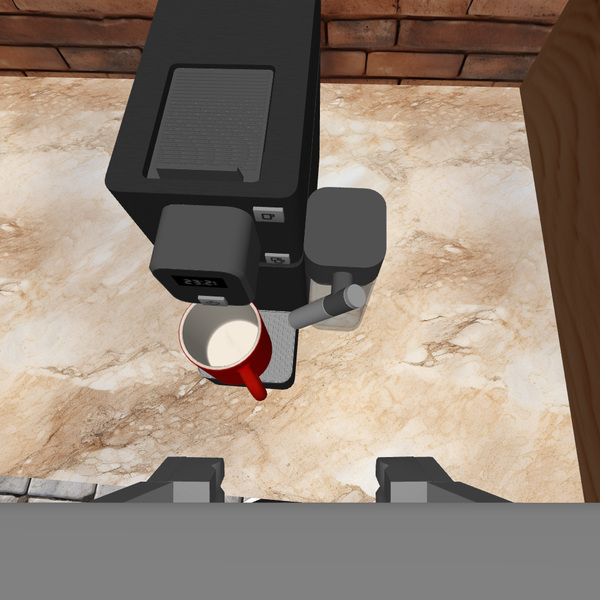}}{Wrist view}}
\begin{casepanel}
\caseheading{Offline plans and closed-loop outcomes}
\noindent\caseimage{.325}{\includegraphics[width=\linewidth]{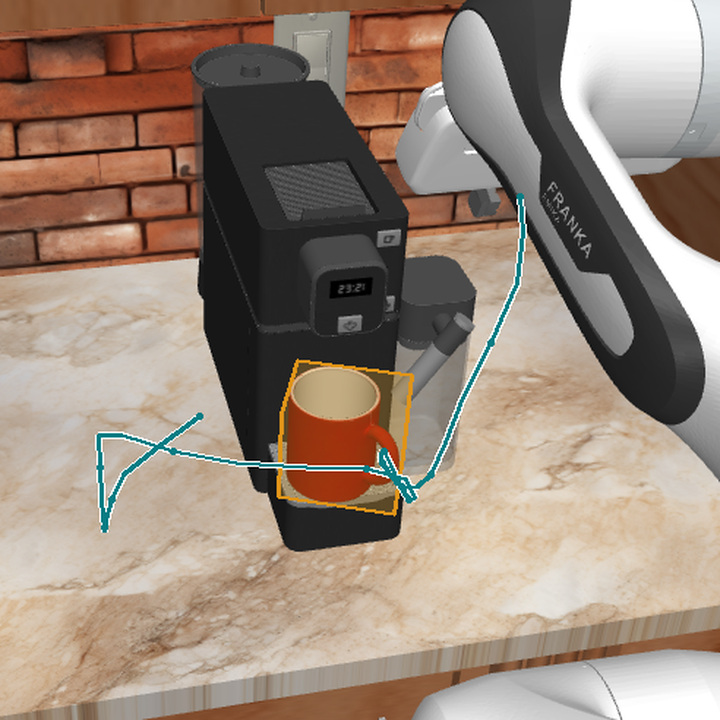}}{\textbf{Astra}\\Offline composite score: 99.53\\Closed-loop outcome: success}\hfill
\caseimage{.325}{\includegraphics[width=\linewidth]{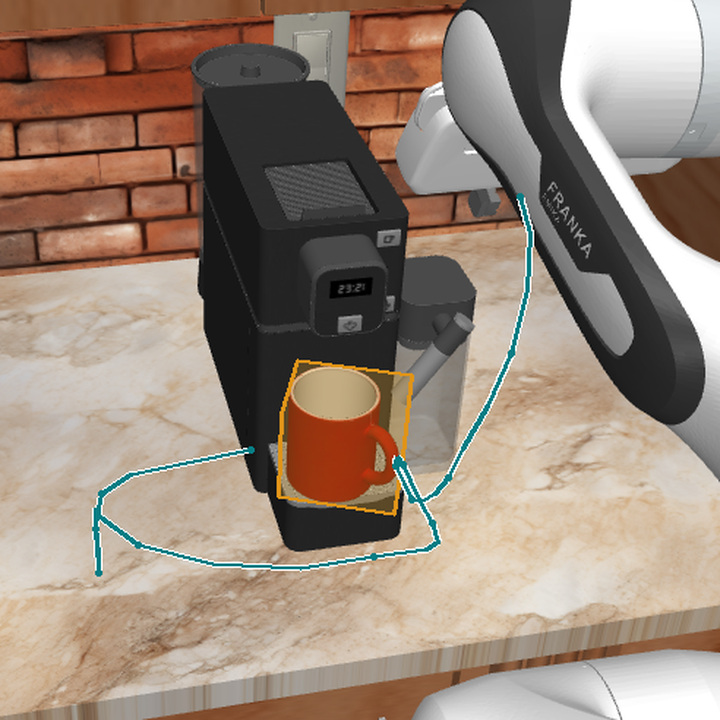}}{\textbf{Sol}\\Offline composite score: 99.41\\Closed-loop outcome: success}\hfill
\caseimage{.325}{\includegraphics[width=\linewidth]{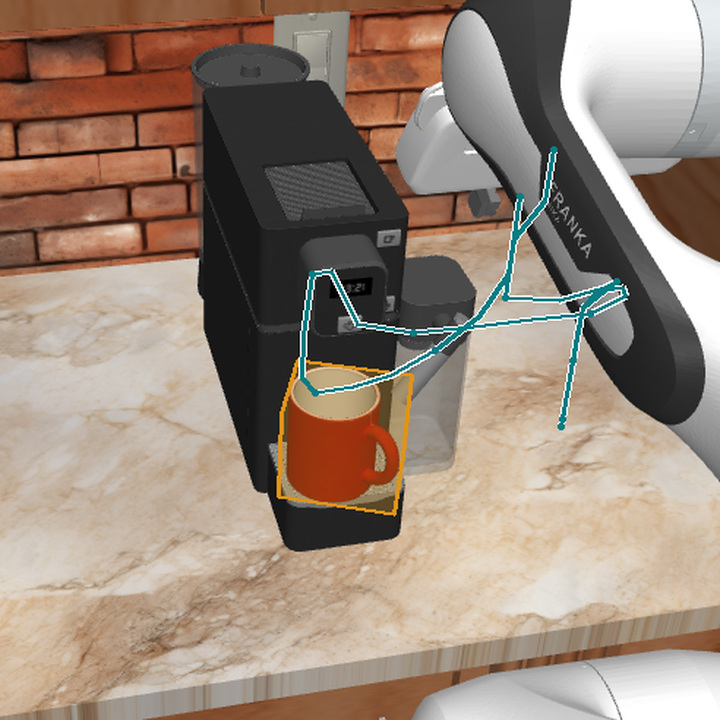}}{\textbf{Luna}\\Offline composite score: 1.35\\Closed-loop outcome: failure}\par
\end{casepanel}
\begin{casepanel}
\caseheading{Recorded simulator observations}
\noindent\caseimage{.325}{\includegraphics[width=\linewidth]{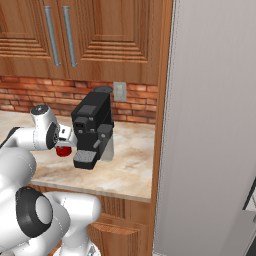}}{Astra, step 255}\hfill
\caseimage{.325}{\includegraphics[width=\linewidth]{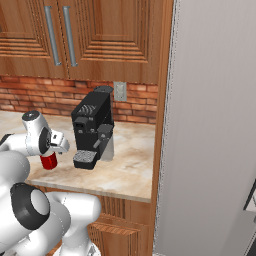}}{Sol, step 243}\hfill
\caseimage{.325}{\includegraphics[width=\linewidth]{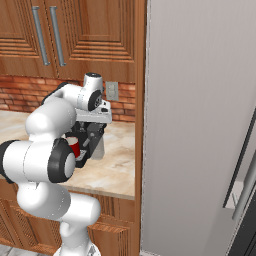}}{Luna, step 450}\par
\captionof{figure}{\textbf{Task-level agreement on the coffee mug.}
The teal lines show the three offline plans relative to the mug's initial region (amber). Astra and Sol satisfied both offline stages and succeeded in simulation at steps 289 and 275, respectively. Luna received a partial grasp score and zero for placement. At the step limit the mug was still under the dispenser.}
\label{fig:case-mug}
\end{casepanel}

\subsection{Transferring an onion from a plate to a pan}
\caseinputs{Pick the onion from the plate and place it in the pan.}%
{\caseimage{.225}{\includegraphics[width=\linewidth]{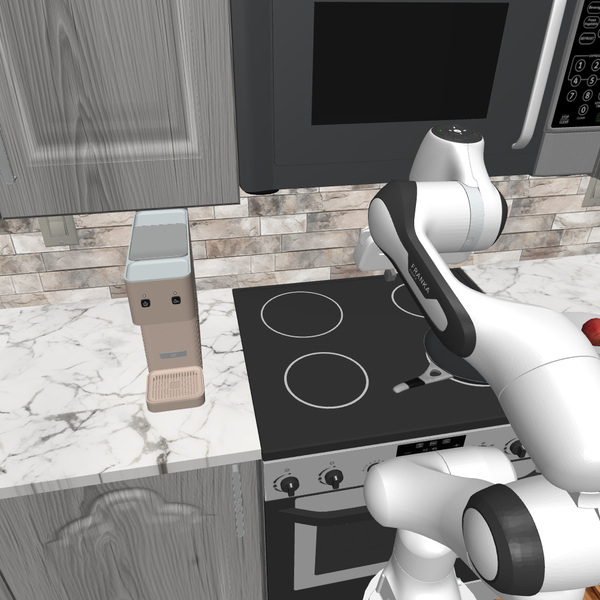}}{Left view}}%
{\caseimage{.225}{\includegraphics[width=\linewidth]{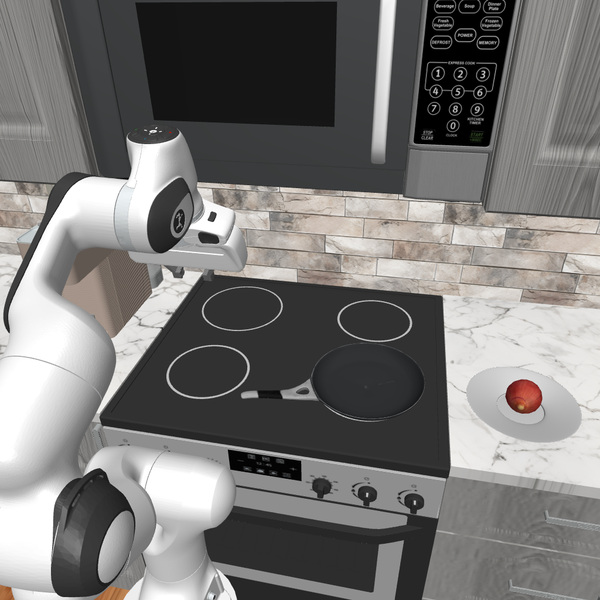}}{Right view}}%
{\caseimage{.225}{\includegraphics[width=\linewidth]{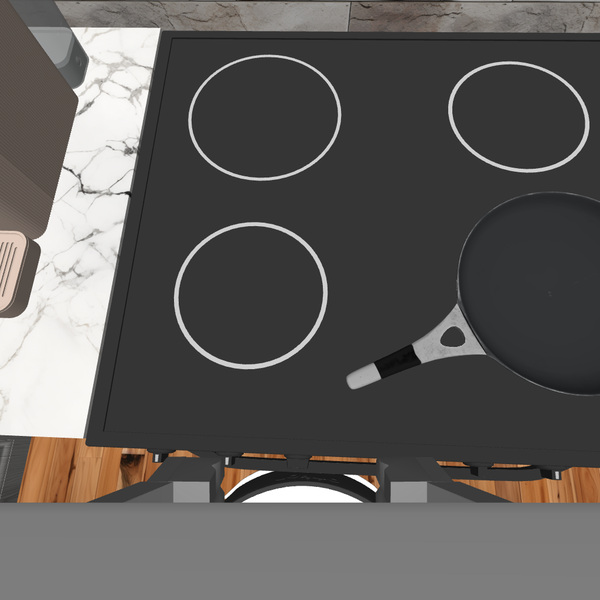}}{Wrist view}}
\begin{casepanel}
\caseheading{Luna's offline prediction and closed-loop observations}
\noindent\caseimage{.325}{\includegraphics[width=\linewidth]{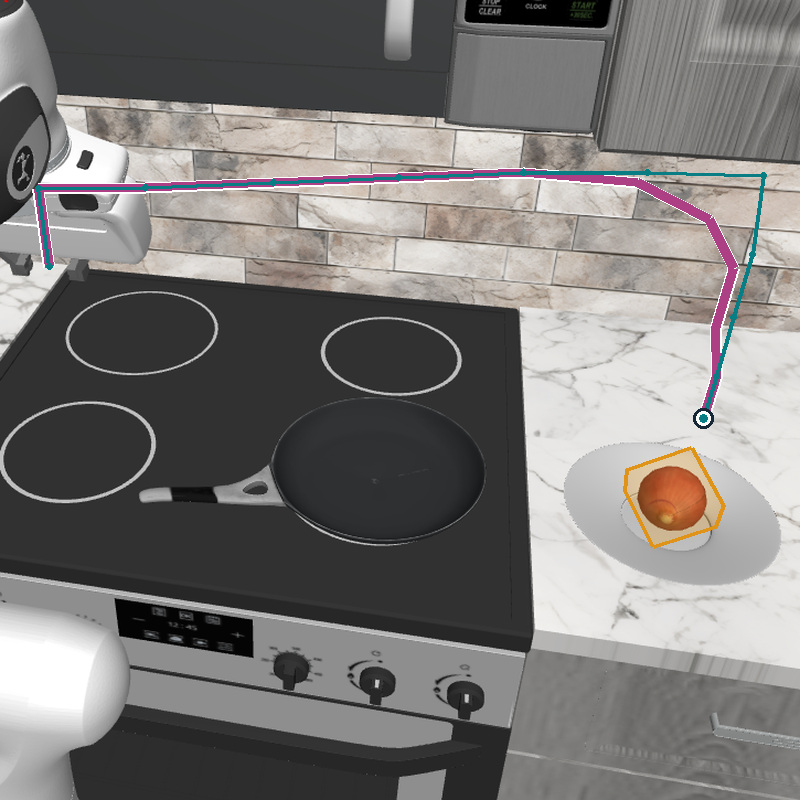}}{Predicted and reconstructed approach}\hfill
\caseimage{.325}{\includegraphics[width=\linewidth]{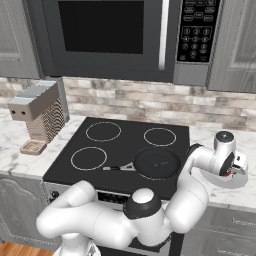}}{Simulation, step 150}\hfill
\caseimage{.325}{\includegraphics[width=\linewidth]{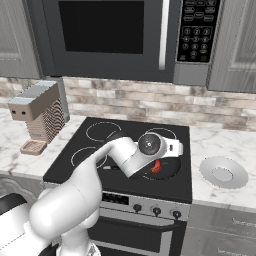}}{Simulation, step 300}\par
\captionof{figure}{\textbf{Successful execution despite a low offline score.}
The ring marks the planned gripper closure outside the onion region (amber). The grasp earns 12.16 and the dependent placement earns zero, giving $R=6.08$. During the approach, the reconstructed TCP (purple) deviates from the predicted TCP (teal) by up to 5.9\,cm at the sampled poses. The grasp-stage tracking score of 16.41 falls below the cutoff of 40, leaving $S=0.61$. The two simulator views show Luna's separate closed-loop run, which completed the transfer at step 324. The standalone $\pi_{0.5}$ executor also succeeded at step 324 with the task instruction alone.}
\label{fig:case-onion}
\end{casepanel}

\subsection{Placing a piece of cake in a mixer bowl}
\caseinputs{Place the cake in the stand mixer bowl.}%
{\caseimage{.175}{\includegraphics[width=\linewidth]{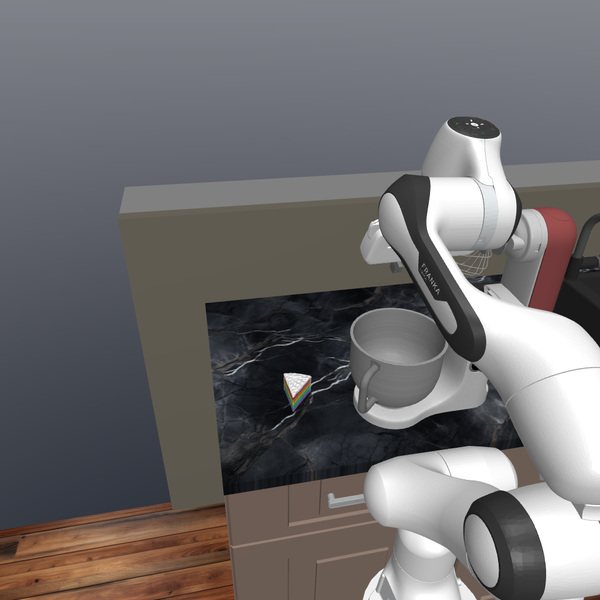}}{Left view}}%
{\caseimage{.175}{\includegraphics[width=\linewidth]{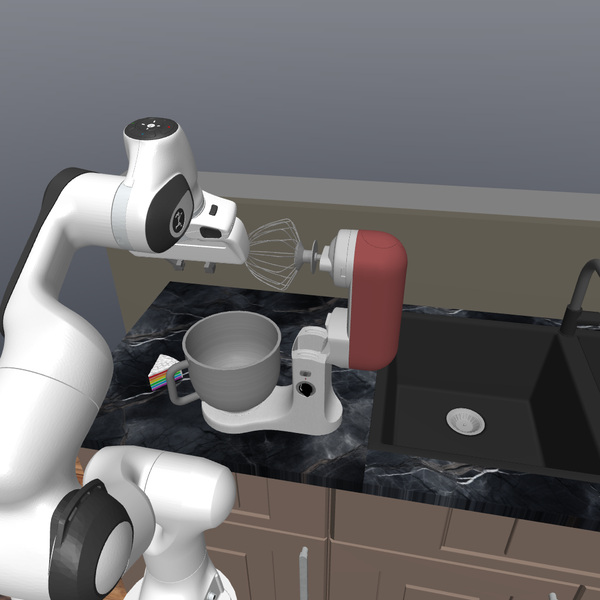}}{Right view}}%
{\caseimage{.175}{\includegraphics[width=\linewidth]{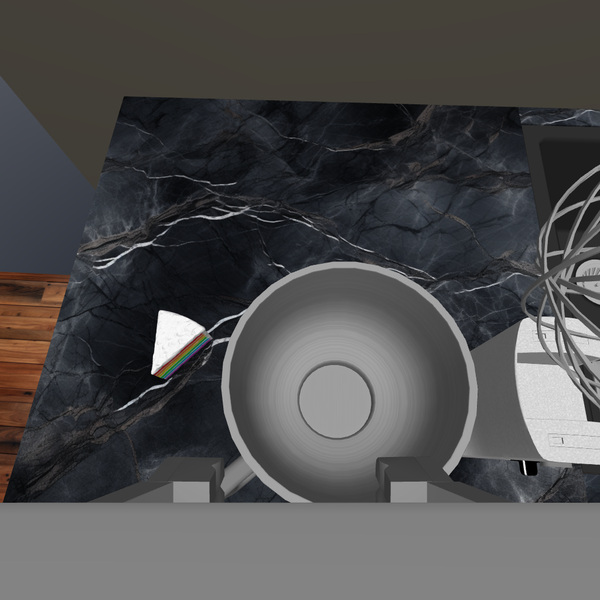}}{Wrist view}}
\begin{casepanel}
\caseheading{From planned placement to observed outcome}
\noindent\caseimage{.238}{\includegraphics[width=\linewidth]{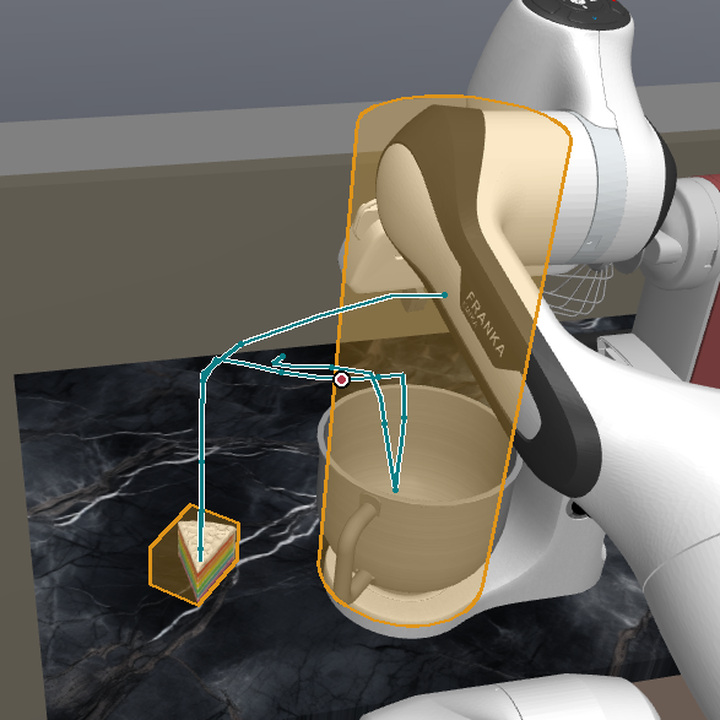}}{Offline plan}%
\hfill\caseimage{.238}{\includegraphics[width=\linewidth]{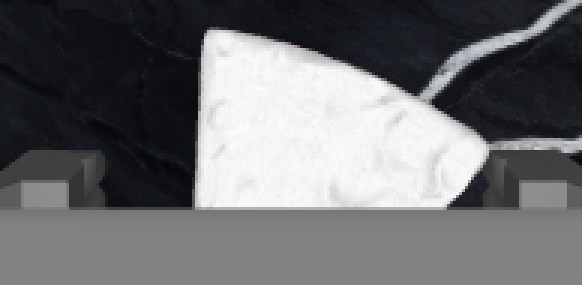}}{Grasp approach\\Step 97}%
\hfill\caseimage{.238}{\includegraphics[width=\linewidth]{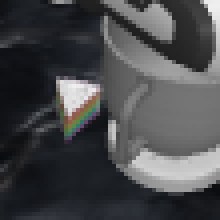}}{Cake beside bowl\\Step 336}%
\hfill\caseimage{.238}{\includegraphics[width=\linewidth]{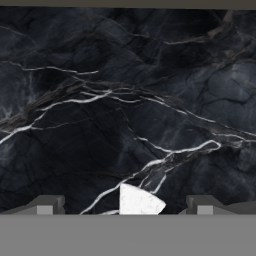}}{Last wrist view\\Step 581}\par
\captionof{figure}{\textbf{A plausible offline placement can still fail during transfer.}
In the offline plan, the amber regions mark the initial cake position and the accepted placement volume. The ring marks the point scored for placement. Astra earns full scores for both stages and an offline composite score of 99.61. In simulation, the cake slipped after the lift. At step 336 it is visible on the counter beside the bowl. It remained outside the bowl when the episode ended at step 600.}
\label{fig:case-cake}
\end{casepanel}

\subsection{Luna's grasps on a drawer and a pan}

\begin{casepanel}
\textbf{Instruction.} \emph{Open the left drawer and place the smaller measuring cup on the counter.}\par\smallskip
\centerline{\includegraphics[width=.78\linewidth]{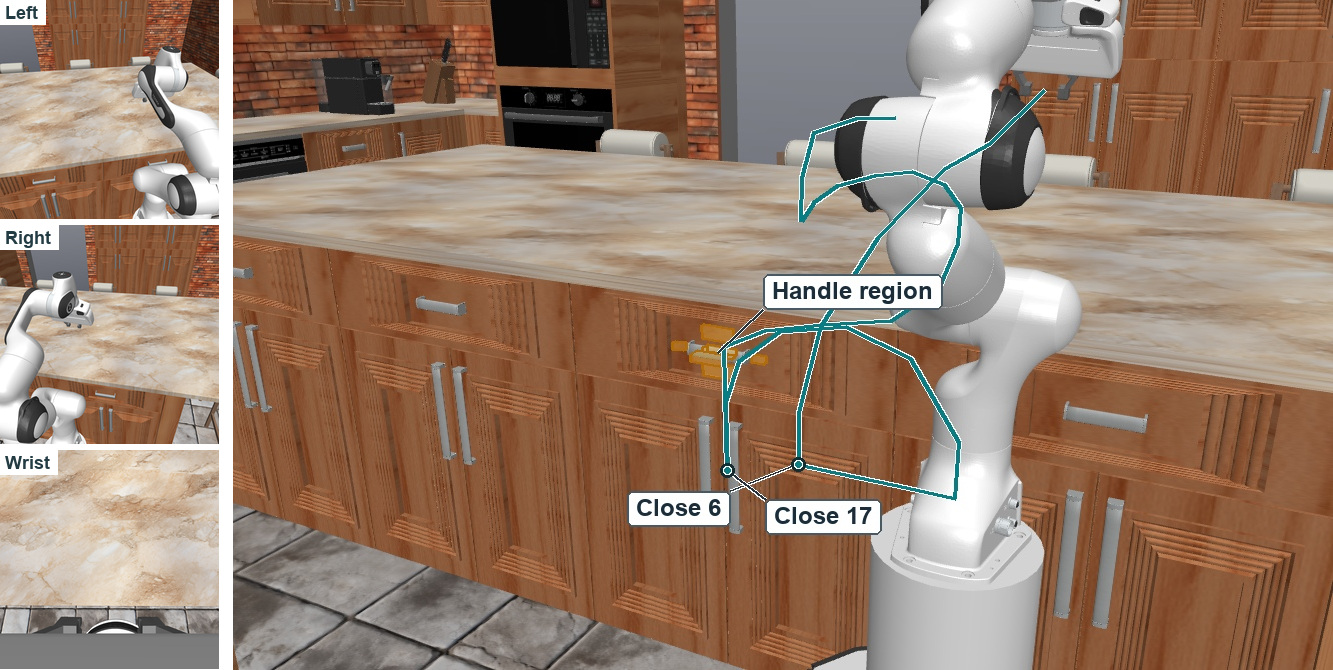}}
\captionof{figure}{\textbf{Both gripper closures miss the drawer handle.}
The three initial observations appear at left. A wider rendering of the same initial scene at right shows Luna's predicted TCP path (teal) and the accepted handle region (amber). The rings mark closures at waypoints 6 and 17, respectively 61.7 and 50.8\,cm from the nearest handle-region center in 3D. Without a grasp on the handle, the opening stage and the dependent retrieval and placement stages all score zero.}
\label{fig:case-luna-drawer}
\end{casepanel}

\begin{casepanel}
\textbf{Instruction.} \emph{Tilt the pan slightly to move the corn inside the pan. Place the pan back down on the stove once finished.}\par\smallskip
\centerline{\includegraphics[width=.78\linewidth]{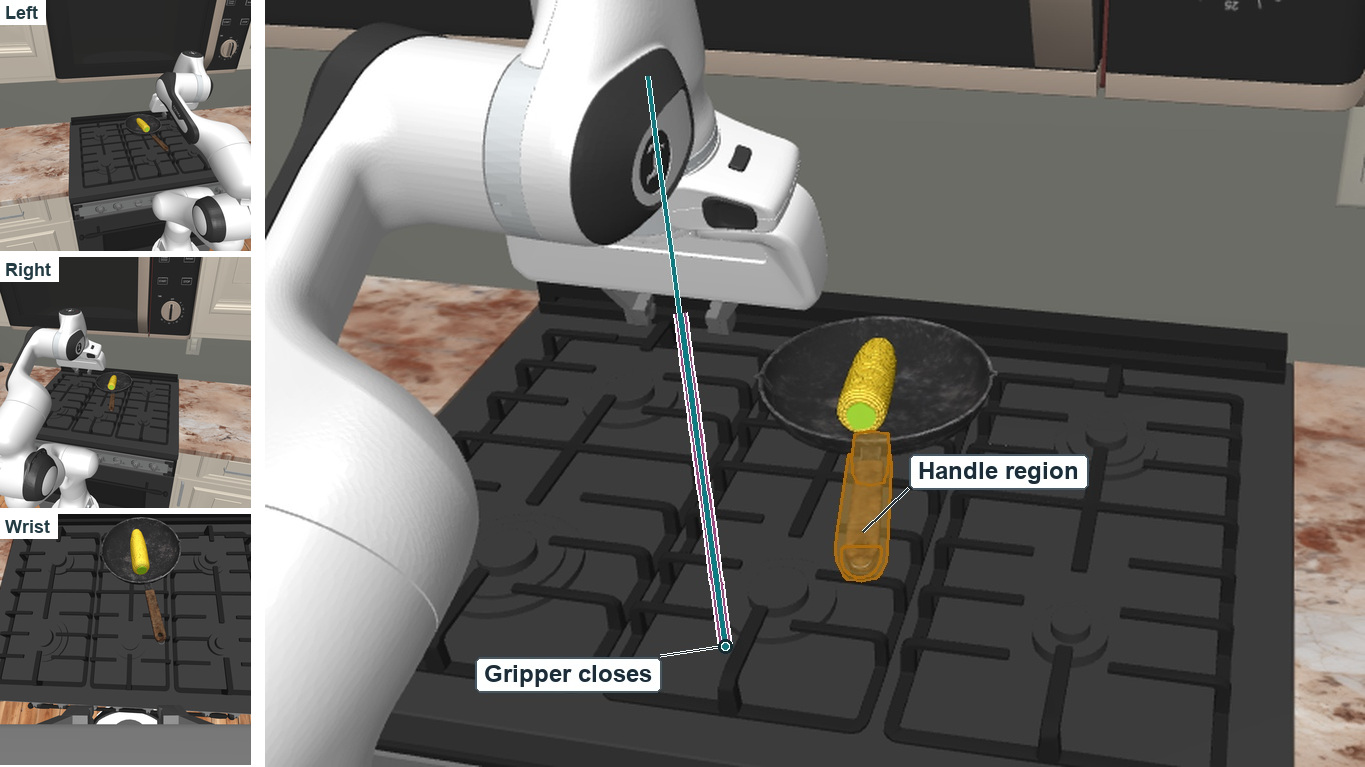}}
\captionof{figure}{\textbf{The arm follows a vertical path beside the pan handle.}
The enlarged right-camera view overlays the predicted TCP path (teal) and the accepted handle region (amber). All 36 waypoints share the same horizontal coordinates. The gripper descends nearly 30\,cm and closes at waypoint 8 (ring). The reconstructed approach (purple) closely follows the prediction ($W_i=99.64$, $C_i=100$), but the grasp earns only 7.95. Tilting and returning the pan then score zero, giving $R=2.65$.}
\label{fig:case-luna-pan}
\end{casepanel}